\documentclass[10pt,twocolumn,letterpaper]{article}

\usepackage[pagenumbers]{cvpr} 

\usepackage[dvipsnames]{xcolor}

\usepackage{multirow}
\usepackage{tabularx}
\usepackage{adjustbox}
\usepackage{array}
\usepackage{kotex}
\usepackage{graphicx}
\usepackage[dvipsnames]{xcolor}
\usepackage{pifont}
\usepackage{tikz}
\usetikzlibrary{tikzmark}

\newcolumntype{R}[2]{%
    >{\adjustbox{angle=#1,lap=\width-(#2)}\bgroup}%
    l%
    <{\egroup}%
}
\newcommand{\nj}[1]{\textcolor{black}{#1}}
\newcommand{\jy}[1]{\textcolor{black}{#1}}

\definecolor{cvprblue}{rgb}{0.21,0.49,0.74}
\definecolor{mgt}{rgb}{0.8, 0.1, 0.8}

\definecolor{apricot}{rgb}{0.94, 0.76, 0.65}
\definecolor{grey}{rgb}{0.9, 0.9, 0.9}
\definecolor{Gray1}{gray}{0.95}
\definecolor{Gray2}{gray}{0.99}
\newcommand{\CC}[1]{\cellcolor{Gray1}}
\newcommand\CCG[1][]{\CC{10}}

\usepackage[pagebackref,breaklinks,colorlinks,citecolor=cvprblue]{hyperref}
\usepackage{colortbl}

\def\paperID{6459} 
\def\confName{CVPR}
\def\confYear{2024}

\title{What, How, and When Should Object Detectors Update \\in Continually Changing Test Domains?}

\author{Jayeon Yoo$^1$~~~~~Dongkwan Lee$^1$~~~~~Inseop Chung$^1$~~~~~Donghyun Kim$^{2*}$~~~~~Nojun Kwak$^{1*}$\\
$^1$Seoul National University~~~~~$^2$Korea University\\
{\tt\small $^1$\{jayeon.yoo, biancco, jis3613, nojunk\}@snu.ac.kr $^2$d\_kim@korea.ac.kr}
}

\begin{document}
\maketitle
\begin{abstract}

It is a well-known fact that the performance of deep learning models deteriorates when they encounter a distribution shift at test time. Test-time adaptation (TTA) algorithms have been proposed to adapt the model online while inferring test data. However, existing research predominantly focuses on classification tasks through the optimization of batch normalization layers or classification heads, but this approach limits its applicability to various model architectures like Transformers and makes it challenging to apply to other tasks, such as object detection. In this paper, we propose a novel online adaption approach for object detection in continually changing test domains, considering which part of the model to update, how to update it, and when to perform the update. By introducing architecture-agnostic and lightweight adaptor modules and only updating these while leaving the pre-trained backbone unchanged, we can rapidly adapt to new test domains in an efficient way and prevent catastrophic forgetting. Furthermore, we present a practical and straightforward class-wise feature aligning method for object detection to resolve domain shifts. Additionally, we enhance efficiency by determining when the model is sufficiently adapted or when additional adaptation is needed due to changes in the test distribution. Our approach surpasses baselines on widely used benchmarks, achieving improvements of up to 4.9\%p and 7.9\%p in mAP for COCO $\rightarrow$ COCO-corrupted and SHIFT, respectively, while maintaining about 20 FPS or higher.
\end{abstract}

\section{Introduction}
\label{sec:intro}


Although deep learning models have demonstrated remarkable success in numerous vision-related tasks, they remain susceptible to domain shifts where the test data distribution differs from that of the training data~\cite{boudiaf2022parameter,TTT,TTT++}. In real-world applications, domain shifts frequently occur at test-time due to natural variations, corruptions, changes in weather conditions \textit{(e.g., fog, rain)}, camera sensor differences \textit{(e.g., pixelate, defocus blur)}, and various other factors. Test-Time Adaptation (TTA)~\cite{boudiaf2022parameter,TTT,TTT++,TENT,EATA,MEMO} has been proposed to solve the domain shifts in test-time by adapting models to a specific target (test) distribution in an online manner. Furthermore, it is essential to take into account continuously changing test distributions, as the test distribution has the potential to undergo changes and developments as time progresses (\ie, Continual Test-time Adaptation (CTA)). For instance, autonomous driving systems may experience transitions from clear and sunny conditions to rainy or from daytime to nighttime, which causes continually changing domain shifts~\cite{shift}. While it is an important research topic, continual test-time adaptation for object detection has not been well explored. 




\begin{figure}[t]
\centerline{\includegraphics[width=0.5\textwidth]{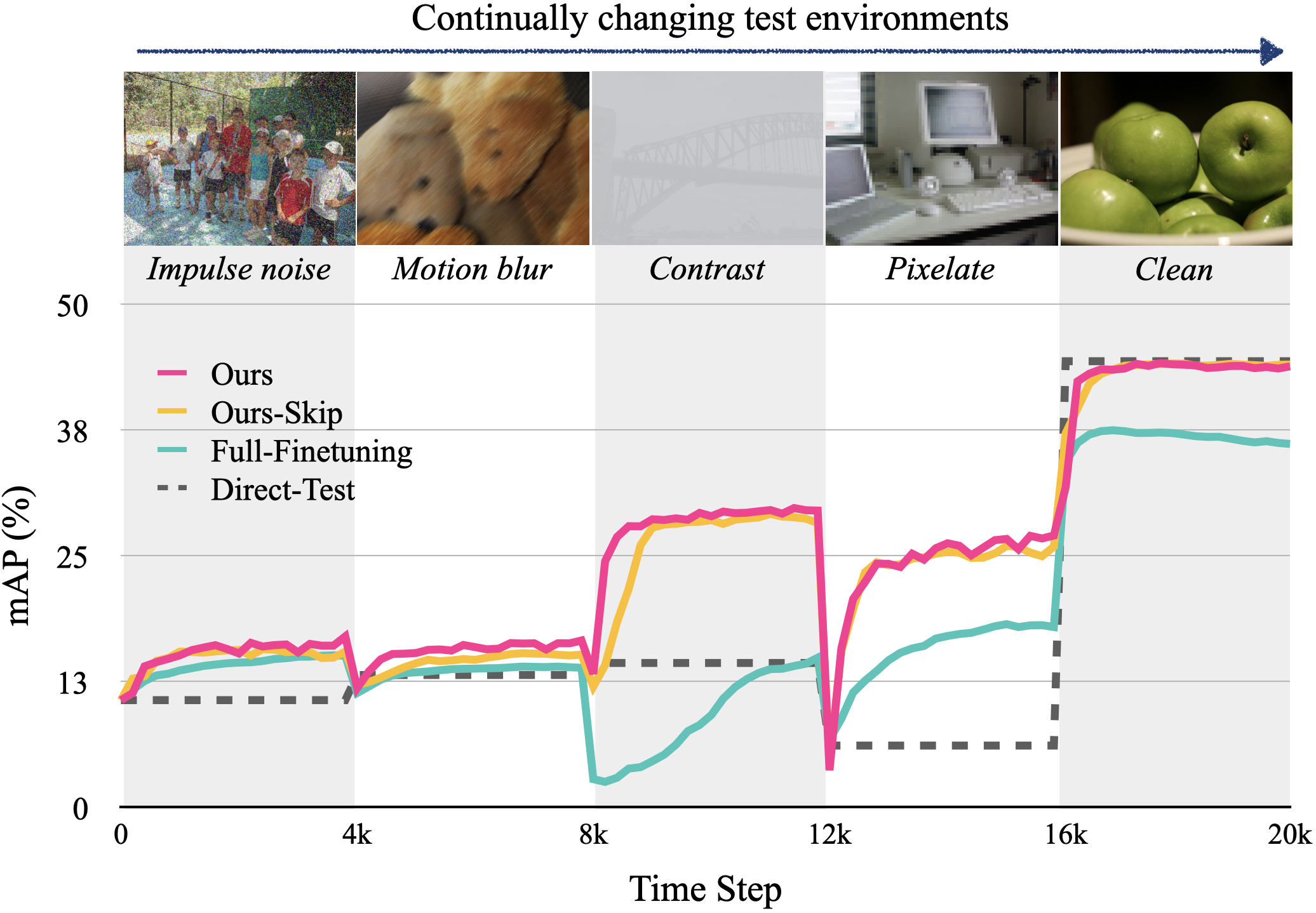}}
\vspace{-2mm}
\caption{We propose an online adaptation method for object detection in continually changing test domains. Object detectors trained with clean images suffer from performance degradation due to various corruption, such as camera sensor degradation or environmental changes \textit{(Direct-Test)}. Updating full parameters for online adaptation require a large number of test samples and vulnerable to drastic domain changes \textit{(Full-Finetuning)}, while using only our lightweight adaptor is robust and quickly adapts within a few time steps \textit{(Ours)}. We can further improve efficiency by skipping unnecessary adaptation steps \textit{(Ours-Skip)}.}
\label{fig:teaser}
\vspace{-2mm}
\end{figure}
%


Recently, several TTA methods ~\cite{ActMAD,TeST,STFAR} tailored for object detection have been proposed. ActMAD~\cite{ActMAD} aligns all the output feature maps ($\mathbb{R}^{C\times H\times W} $) after Batch Normalization (BN) layers~\cite{Batchnorm} to adapt the test domain to be similar to that of the training domain. However, this approach requires significant memory during adaptation and does not explicitly consider the objects present in the image. TeST~\cite{TeST} and STFAR~\cite{STFAR} adapt to a test domain by utilizing weak and strong augmented test samples with a teacher-student network~\cite{sohn2020fixmatch}, but they significantly increase inference costs since they require
additional forward passes and update steps. Also, these methods update all network parameters, making them highly inefficient in online adaptation and vulnerable to losing task-specific knowledge when the test domain experiences continual or drastic changes.

In this paper, we aim to develop an efficient continual test-time adaptation (CTA) method for object detection. We investigate the following three key aspects to improve efficiency; \textbf{\textit{what to update}}: while previous TTA methods for object detection~\cite{ActMAD,TeST,STFAR} use full fine-tuning, updating all parameters at test time, they are inefficient and prone to losing task-specific knowledge in relatively complex object detection tasks. Updating BN layers, as done in many TTA methods for classification~\cite{TENT,EATA,MEMO,CAFA}, is not as effective for object detection, given its smaller batch size compared to classification and the limitation in applying various backbones, such as Transformer~\cite{liu2021swin, vaswani2017attention}.
\textbf{\textit{how to update}}: 
several previous TTA methods for object detection~\cite{TeST,STFAR} adapt the model by using teacher-student networks, resulting in a significant decrease in inference speed, which is detrimental during test time. While another existing method~\cite{ActMAD} aligns feature distributions for adaptation, it does not consider each object individually, focusing only on image features, making it less effective for object detection.
\textbf{\textit{when to update}}: most TTA or CTA methods update models using all incoming test samples. However, it is inefficient to update continuously the model if it is already sufficiently adapted when the change of the test domain is not significant.

To this end, (1) we propose an efficient continual test-time adaptation method for object detectors to adapt to continually changing test domains through the use of lightweight adaptors which require only
0.54\%$\sim$0.89\% additional parameters compared to the full model.
It exhibits efficiency in parameters, memory usage, and adaptation time, along with robustness to continuous domain shifts without catastrophic forgetting. Additionally, it demonstrates wide applicability to various backbone types compared to BN-based TTA methods~\cite{TENT,EATA,MEMO,CAFA,TTN,DELTA}. (2) To enhance the adaptation effectiveness in the object detection task, we align the feature distribution of the test domain with that of the training domain at both the image-level and object-level using only the mean and variance of features. For estimating the mean of the test domain features, we employ Exponentially Moving Average (EMA) as we can leverage only the current incoming test samples, not the entire test domain data. Due to the unavailability of training data access, we utilize only the mean and variance of the features from a few training samples.
(3) We also introduce two novel criteria that do not require additional resources to determine when the model needs adaptation to enhance efficiency in a continually changing test domain environment. As illustrated in Fig.~\ref{fig:teaser}, our approach \textit{Ours}, employing adaptors, tends to adapt much faster to domain changes compared to full parameter updates. This enables efficient TTA by using only a few test samples to update the adaptor and skipping the rest of the updates as shown in \textit{Ours-Skip}.



Our main contributions are summarized as follows:

\begin{itemize}
\item We introduce an architecture-agnostic lightweight adaptor, constituting only a maximum of 0.89\% of the total model parameters, into the backbone of the object detector to adapt the model in a continually changing test domain. 
This approach ensures efficiency in parameters, memory usage, and adaptation speed, demonstrating the robust preservation of task-specific knowledge owing to its inherent structural characteristics.
\item We propose a straightforward and effective adaptation loss for CTA in object detection tasks. This is achieved by aligning the distribution of training and test domain features at both the image and object levels, utilizing only the mean and variance of a few training samples and EMA-updated mean features of the test domain.
\item We also propose two criteria to determine when the model requires adaptation, enabling dynamic skipping or resuming adaptation as needed. This enhancement significantly boosts inference speed by up to about 2 times while maintaining adaptation performance.

\item Our adaptation method proves effective for diverse types of domain shifts, including weather changes and sensor variations, regardless of whether the domain shift is drastic or continuous. In particular, our approach consistently improves the mAP by up to 7.9\% in COCO$\rightarrow$COCO-C and SHIFT-Discrete/Continuous with higher than 20 FPS.
\end{itemize}

\section{Related Work}
\label{sec:relwork}


\noindent\textbf{Test-time adaptation.}
Recently, there has been a surge of interest in research that adapts models online using unlabeled test samples while simultaneously inferring the test sample to address the domain shift problem, where the test data distribution differs from that of the training data. 
There are two lines for online adaptation to the test domain, \textit{Test-time Training (TTT)} and \textit{Test-time Adaptation (TTA)}. TTT~\cite{TTT,TTT++,Ttaps,Mt3} involves modifying the model architecture during training to train it with self-supervised loss, allowing adaptation to the test domain in the test time by applying this self-supervised loss to the unlabeled test samples. On the other hand, TTA aims to adapt the trained model directly to the test domain without specifically tailored model architectures or losses during training time.
NORM~\cite{NORM} and DUA~\cite{DUA} address the domain shifts by adjusting the statistics of batch normalization (BN) layers using the current test samples, without updating other parameters, inspired by~\cite{li2017adabn}. Following this, \cite{TENT, EATA, DELTA, TTN} and \cite{CAFA} update the affine parameters of BN layers using unsupervised loss, entropy minimization loss to enhance the confidence of test data predictions, and feature distribution alignments loss, respectively. Several studies~\cite{iwasawa2021t3a,TAST} update the classifier head using the pseudo-prototypes from the test domain. However, these methods limit their applicability to architectures without BN layers or to object detection tasks that involve multiple objects in a single image. Others~\cite{MEMO, ActMAD, TTAC} update full parameters for online adaptation to the test domain in an online manner, but this approach is inefficient and susceptible to the noisy signal from the unsupervised loss. While existing TTA methods are oriented towards classification tasks, we aim to propose an effective and efficient method for online adaptation in the object detection task.
\noindent\textbf{Continual test-time adaptation.} Recent studies~\cite{SAR,CoTTA} point out that existing TTA methods have primarily focused on adapting to test domains following an i.i.d assumption and may not perform well when the test data distribution deviates from this assumption. \cite{CoTTA} introduces a \textit{Continual TTA (CTA)} method designed for scenarios where the test domain continuously changes over time. This poses challenges in preventing the model from over-adapting to a particular domain shift and preserving the knowledge of the pre-trained model to avoid catastrophic forgetting.
In the field of CTA, the self-training strategy adopting an Exponentially Moving Average (EMA) teacher-student structure is attracting interest as an effective algorithm enabling robust representation to be learned through self-knowledge distillation. In many studies, the EMA teacher-student structure and catastrophic restoration of source model weights have been proposed as a solution to achieve the goal of CTA \cite{CoTTA, xiao2023energy, PETAL}. Approaches using source replay \cite{RMT}, and anti-forgetting regularization \cite{EATA} have also achieved good performances in robust continuous adaptation. Furthermore, there is growing attention on methods that mitigate the computational and memory challenges associated with CTA, such as \cite{MECTA}, which relies on updates to batch normalization statistics.

\noindent\textbf{Test-time adaptive object detection.} Research on \textit{TTA for Object Detection (TTAOD)} is progressively emerging~\cite{ActMAD, TeST, STFAR, MemCLR}. Most existing TTAOD methods~\cite{TeST, STFAR, MemCLR} exploit a teacher-student network to adapt to the test domain, following the self-training approach commonly employed in Unsupervised Domain Adaptation for object detection~\cite{RoyChowdhury_2019_CVPR, deng2021unbiased, khodabandeh2019robust, kim2019self}. However, it is inefficient for TTA due to data augmentation requirements and additional forward and backward steps, resulting in slower inference speeds and higher memory usage. Another approach, ActMAD~\cite{ActMAD}, aligns the distributions of output feature maps after all BN layers along the height, width, and channel axes to adapt to the test domain. However, this location-aware feature alignment is limited to datasets with fixed location priors, such as driving datasets, and is less effective for natural images like COCO. Additionally, \textit{CTA for Object Detection (CTAOD)} have not been thoroughly explored. Therefore, there is a need for an effective CTAOD method considering memory and time efficiency.

\section{Method}
To enable the efficient and effective Continual Test-time Adaptation of Object Detectors (CTAOD), we introduce an approach that specifies \textit{which part of the model should be updated}, describes \textit{how to update} those using unlabeled test data, and determines \textit{whether we perform model updates or not} to improve efficiency.

\subsection{Preliminary}
Assume that we have an object detector $h\circ g_{\Theta}$, here $h$ and $g$ are the RoI head and the backbone, respectively with their parameters being $\Theta$. The training dataset is denoted as $D_{train}=\{(x_i, y_i)\}_{i=1}^{N}$, where $x_i\sim P_{train}(x)$ and $y_i = (\mathit{bbox}_i, c_i)$, containing information on the bounding box (bbox) and class label $c_i\in \mathcal{C}$. Consider deploying the detector to the test environments where the test data at period $T$ is denoted as $x^{T}_j \sim P^{T}_{test}(x)$, $P^{T}_{test} \neq P_{train}$ and $P^{T}_{test}$ deviates from the i.i.d. assumption. In addition, the domain of $P^{T}_{test}$ continually changes according to $T$ (\ie, $P^{T}_{test} \neq P^{T-1}_{test}$). Our goal is to adapt the detector $h\circ g$ to $P^{T}_{test}$ using only test data $x^{T}_j$ while making predictions.

\subsection{What to update: Adaptation via an adaptor}
\label{subsec:adaptor}
\begin{figure}
\begin{subfigure}{.25\textwidth}
  \centering
  \includegraphics[width=.9\linewidth]{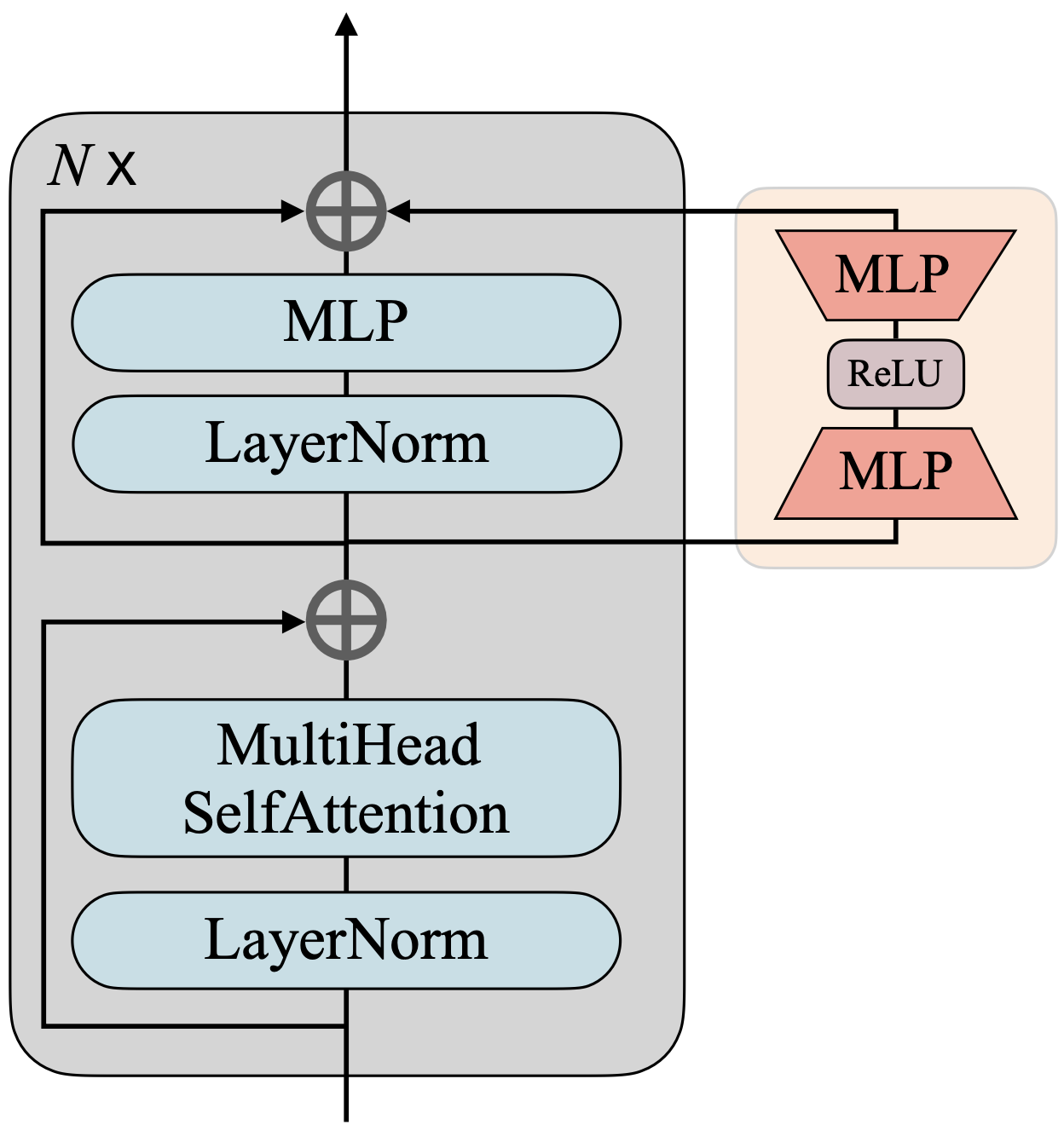}
  \caption{A block of Transformer}
  \label{fig:adaptor_transformer}
\end{subfigure}%
\begin{subfigure}{.25\textwidth}
  \centering
  \includegraphics[width=.9\linewidth]{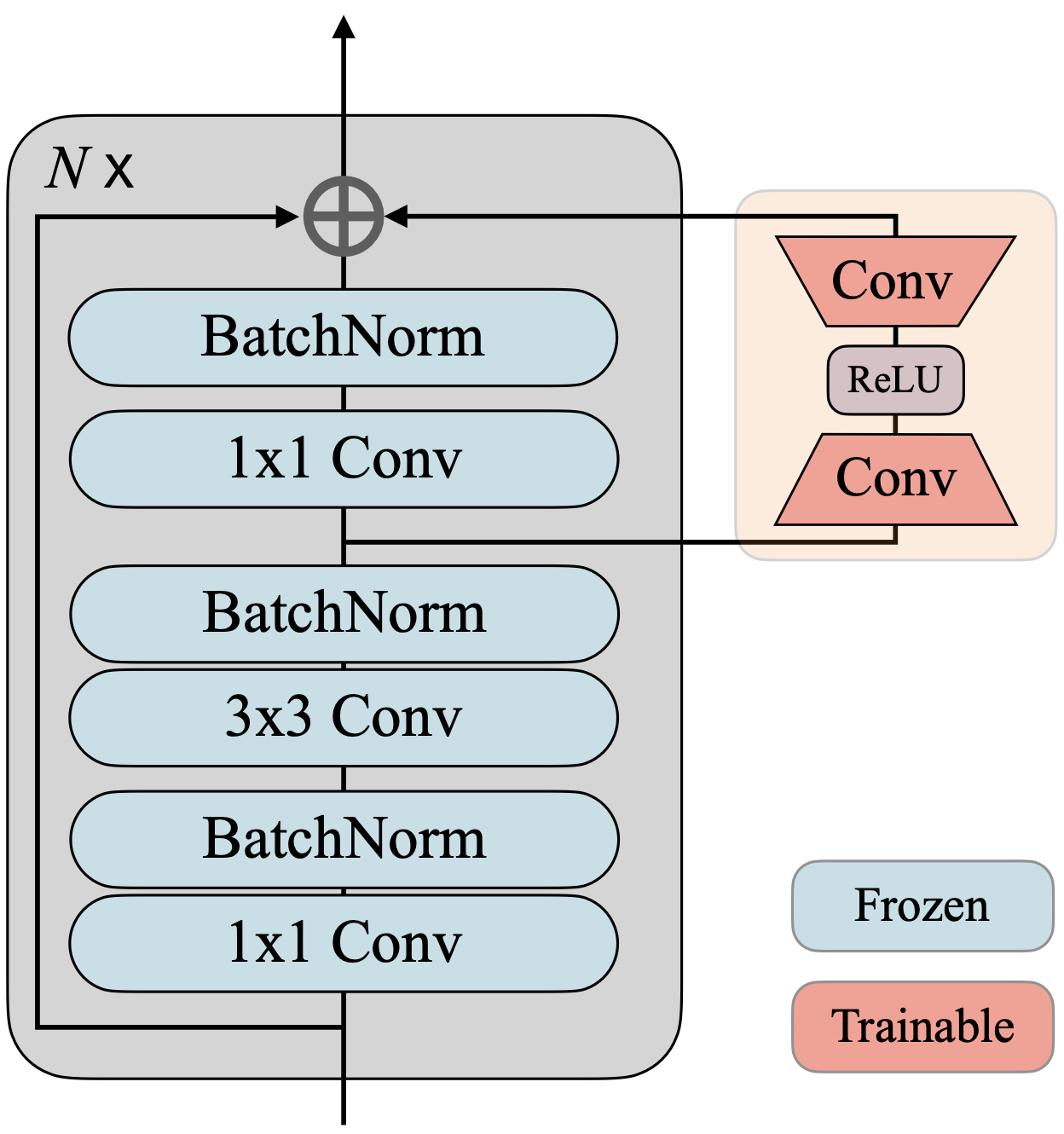}
  \caption{A block of ResNet}
  \label{fig:adaptor_resnet}
\end{subfigure}
\vspace{-2mm}
\caption{We attach an \textcolor{apricot}{adaptor}, which is a shallow and low-rank MLP or CNN, to every $N$ block in parallel. We update only these adaptors while other parameters are frozen. Our approach can be applied to diverse architectures including CNNs and Transformers.}
\vspace{-2mm}
\label{fig:adaptors}
\end{figure}

Previous methods \cite{ActMAD,TeST,STFAR,MemCLR} adapt the model to the test domain by updating all parameters $\Theta$, leading to inefficiency at test time and a high risk of losing task knowledge from the training data. In contrast, we adapt the model by introducing an adaptor with an extremely small set of parameters and updating only this module while freezing $\Theta$. We introduce a shallow adaptor in parallel for each block, inspired by \cite{chen2022adaptformer,hu2021lora}, where transformer-based models are fine-tuned for downstream tasks through parameter-efficient adaptors, as shown in Fig.~\ref{fig:adaptors}. Each adaptor consists of down-projection layers $\mathcal{W}_{down}\in \mathbb{R}^{d\times \frac{d}{r}}$, up-projection layers $\mathcal{W}_{up}\in \mathbb{R}^{\frac{d}{r}\times d}$ and ReLUs, where $d$ denotes the input channel dimension and $r$ is the channel reduction ratio set to 32 for all adaptors.
We use MLP layers for the Transformer block (Fig.~\ref{fig:adaptor_transformer}) and 1$\times$1 convolutional layers for the ResNet block (Fig.~\ref{fig:adaptor_resnet}) to introduce architecture-agnostic adaptors. The up-projection layer is initialized to 0 values so that the adaptor does not modify the output of the block, but as the adaptor is gradually updated, it adjusts the output of the block to adapt to the test domain. Even as the adaptor is updated in the test domain, the original backbone parameter $\Theta$ remains frozen and fully preserved. This structural preservation, as evident in \textit{Ours} in Fig.~\ref{fig:teaser}, enables robust and efficient adaptation to domain changes by maintaining relatively complex task knowledge in object detection and updating very few parameters.

\newcommand{\argmax}{\arg\!\max}
\subsection{How to update: EMA feature alignment}
\label{subsec:loss}
To adapt the object detector to the test domain, we align the feature distribution of the test domain with that of the training data, inspired by \cite{CAFA, TTAC, ActMAD}. In contrast to these methods that solely align image feature distribution, we additionally align object-level features in a class-wise manner, considering class frequency, to enhance its effectiveness for object detection. As the training data is not accessible during test time, we pre-compute the first and second-order statistics, denoted as $\mu_{tr}=\mathbb{E}[F_{tr}]$ and $\Sigma_{tr}=\mathbb{V}\mathrm{ar}[F_{tr}]$, where the operators $\mathbb{E}$ and $\mathbb{V}\mathrm{ar}$ represent the mean and variance respectively. The features $F_{tr}=\{g_{\Theta}(x_{tr})\}$ are computed using only 2,000 training samples, a small subset of the training data. Since a sufficient amount of test domain data is not available at once, and only the current incoming test data, whose features are denoted as $F_{te}^t$, is accessible at time step $t$, we estimate the mean of test data features using an exponentially moving average (EMA) as follows:
\begin{equation}
    \begin{gathered}
        \mu_{te}^t = (1 - \alpha) \cdot \mu_{te}^{t-1} + \alpha \cdot \mathbb{E}[F_{te}^t],\ \ \ \ \text{s.t.}\ \  \mu_{te}^0 = \mu_{tr}.
    \end{gathered}
\label{eq:ema_mean}
\end{equation}
Considering the typically small batch size in object detection compared to classification, we approximate the variance of the test features as $\Sigma_{te}\simeq\Sigma_{tr}$ to reduce instability.

\noindent\textbf{Image-level feature alignment.} We estimate the training and test feature distributions as normal distributions and minimize the KL divergence between them as follows:
\begin{equation}
    \begin{gathered}
        L_{img} = \mathrm{D}_{KL}(\mathcal{N}(\mu_{tr}, \Sigma_{tr}), \mathcal{N}(\mu_{te}^t, \Sigma_{tr})).
    \end{gathered}
\label{eq:kldiv}
\end{equation}

\noindent\textbf{Region-level class-wise feature alignment.} In object detection, we deal with multiple objects within a single image, making it challenging to apply the class-wise feature alignment proposed in~\cite{TTAC}, a TTA method for classification. 
To handle region-level features that correspond to an object, we use ground truth bounding boxes for the training data and utilize the class predictions of RoI pooled features, $f_{te}^t$, for unlabeled test data. 
In object detection, domain shifts often result in lower recall rates, as a significant number of proposals are predicted as background~\cite{SED}. To mitigate this issue, we filter out features with background scores exceeding a specific threshold. Subsequently, we assign them to the foreground class with the highest probability, as follows:
\begin{equation}
    \begin{gathered}
        F_{te}^{k,t} = \{f_{te}^t|\argmax_c{p_{fg}} = k, p_{bg} < 0.5\}, \\
        where\ \ h_{cls}(f_{te}^t) = [p_{fg}, p_{bg}] = [p_0, ..., p_{C-1}, p_{bg}].
    \end{gathered}
\label{eq:assigning}
\end{equation}
We estimate the class-wise feature distribution of the test domain by exploiting $F_{te}^{k,t}$ and Eq.\ref{eq:ema_mean}. Furthermore, we introduce a weighting scheme for aligning features of less frequently appearing classes, taking into account the severe class imbalance where specific instance (\textit{e.g., person}) may appear multiple times within a single image, as follows:
\begin{equation}
    \begin{gathered}
        N^{k,t} = N^{k,t-1} + ||F_{te}^{k,t}||,\ \ \ \text{s.t.}\ \ N^{k,0}=0 \\
        w^{k,t} = \log \left(\frac{\max_i{N^{i,t}}}{N^{k,t}}\right) + 0.01 \\  
        L_{obj} = \sum_k {w^{k,t}}\cdot D_{KL}(\mathcal{N}(\mu_{tr}^k, \Sigma_{tr}^k), \mathcal{N}(\mu_{te}^{k,t}, \Sigma_{tr}^k)).
    \end{gathered}
\label{eq:obj_kldiv}
\end{equation}
Here, the class-wise mean $\mu^k$ and variance $\Sigma^k$ of the training and test data are obtained in the same way as the image-level features.
We can effectively adapt the object detector by updating the model to align the feature distribution at both the image and object levels as $L=L_{img} + L_{obj}$.

\subsection{When to update: Adaptation on demand}
\label{subsec:when}
\begin{figure}
\begin{subfigure}[h]{.45\textwidth}
  \centering
  \includegraphics[width=1.\linewidth]{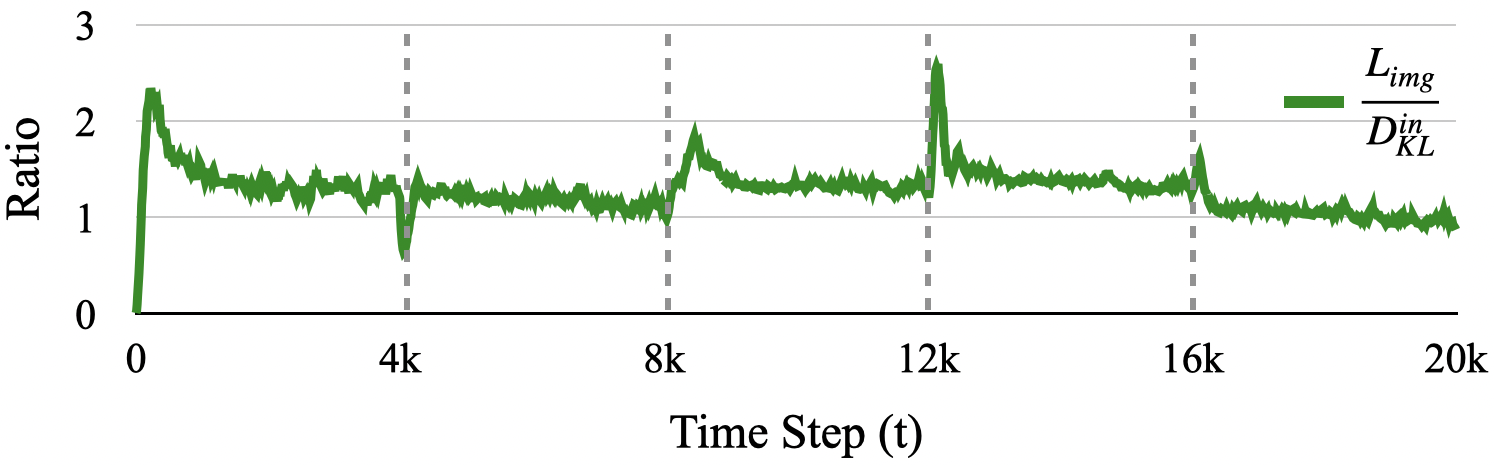}
  \caption{The ratio of $L_{img}$ to $D_{KL}^{in}$}
  \label{fig:skip1}
\end{subfigure}%
\hfill
\begin{subfigure}[h]{.45\textwidth}
  \centering
  \includegraphics[width=1.\linewidth]{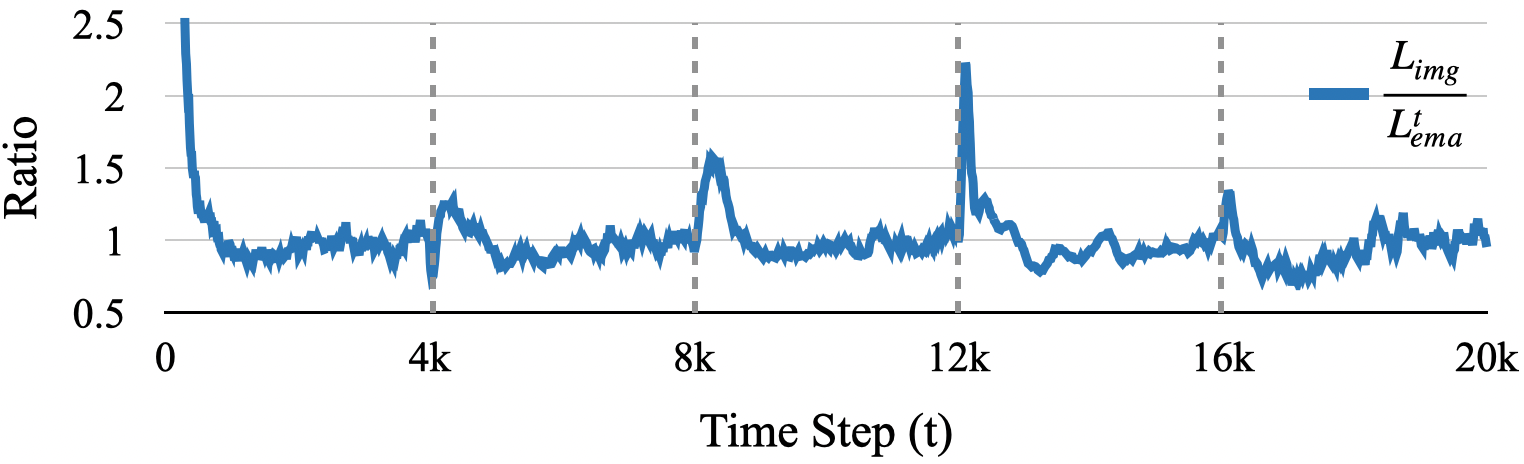}
  \caption{The ratio of $L_{img}$ to $L_{ema}^t$}
  \label{fig:skip2}
\end{subfigure}
\vspace{-2mm}
\caption{The test domain undergoes a shift every 4,000 time steps, and each metric reaches its peak at the same intervals.}
\label{fig:skipping_criteria}
\vspace{-2mm}
\end{figure}

As shown in Fig.~\ref{fig:teaser}, \textit{Ours}, which only updates the adaptor proposed in Sec.~\ref{subsec:adaptor}, efficiently adapts to changes in the test domain, even with a small subset of early test samples. We leverage its rapid adaptation characteristics to reduce computational costs by skipping model updates (\ie, skipping backward passes) when the model has already sufficiently adapted to the current test domain and resuming model updates when confronted with a new test domain. Therefore, we introduce two criteria to determine when to update the model or not as follows: 


\noindent(\textit{Criterion 1}) \textbf{When the distribution gap exceeds the in-domain distribution gap. } Recall that $L_{img}$ (Eq.~\ref{eq:kldiv}) measures the distribution gap between the test and train distributions. We assume a model is well-adapted to the current test domain when $L_{img}$ is closer to the in-domain distribution gap. We measure the in-domain distribution gap by sampling two disjoint subsets, $x_i$ and $x_j$, of training features $F_{tr}$ from Sec.~\ref{subsec:loss} as follows: 
\begin{equation}
D_{KL}^{in} = D_{KL}(\mathcal{N}(\mu_{tr}^i, \Sigma_{tr}^i), \mathcal{N}(\mu_{tr}^{j}, \Sigma_{tr}^j)),
\end{equation}
where $\mu_{tr}^i, \Sigma_{tr}^i$ are obtained from $x_i\sim P_{train}(x)$ and $\mu_{tr}^j, \Sigma_{tr}^j$ from $ x_j\sim P_{train}(x)$. In other words, if $L_{img}$ is noticeably higher than the in-domain distribution gap $D_{KL}^{in}$, we consider a model encountering a test domain whose distribution differs from $P_{train}(x)$ and needs to be updated. Based on this, we introduce a new index $\frac{L_{img}}{D_{KL}^{in}}$. Fig.~\ref{fig:skip1} plots the trend of this index during the model adaptation to a continually changing test domain. It shows that the index has a large value in the early stages of a domain change, decreases rapidly, and then maintains a value close to 1. This index exhibits a similar trend regardless of the backbone type and dataset, as included in the appendix. Therefore, we establish the criterion that model updates are necessary when this index exceeds a certain threshold, $\tau_1$, as $\frac{L_{img}}{D_{KL}^{in}} > \tau_1$.

\noindent(\textit{Criterion 2}) \textbf{When the distribution gap suddenly increases. }
Additionally, we can determine when the test distribution changes and model updates are necessary by observing the trend of the distribution gap (\ie, $L_{img}$). The convergence of $L_{img}$ indicates that a model is well-adapted to the current test domain. To put it differently, $L_{img}$ will exhibit a sudden increase when the model encounters a new test domain. We introduce an additional index, denoted as $\frac{L_{img}}{L_{ema}^t}$, representing the ratio of the current $L_{img}$ to its exponentially moving average $L_{ema}^t$ at time $t$. We calculate it using the following formula: $L_{ema}^t = 0.99\cdot L_{ema}^{t-1} + 0.01 \cdot L_{img}$. Fig.~\ref{fig:skip2} illustrates the trend of the ratio of $L_{img}$ over the timesteps. It tends to reach a value of 1 as the loss stabilizes at a specific level. Nevertheless, when the model encounters shifts in the test distribution, the ratio experiences a sharp increase, indicating the necessity of a model update when it exceeds a specific threshold, $\tau_2$, as $\frac{L_{img}}{L_{ema}^t} > \tau_2$.

\noindent If at least one of the two criteria is satisfied, we conclude that the model requires adaptation and proceed to update it.

\section{Experiments}
\nj{Sec.~\ref{sec:datasets} presents the two object detection benchmark datasets with \jy{test distributions that change continuously, either in a drastic or gradual manner,} and our} implementation detail is in~\ref{sec:implementation}. \nj{Sec.~\ref{sec:results} compares our method with other TTA baselines described in Secs.~\ref{sec:baselines}.}. We present detailed ablation studies of our method \nj{analyzing} the effectiveness and efficiency of our method in terms of what, how, and when to update the models for CTAOD in Sec.~\ref{sec:ablation_study}.
\subsection{Datasets}
\label{sec:datasets}
We experiment with the following three scenarios. 

\newcolumntype{A}{>{\centering}p{0.015\textwidth}}
\newcolumntype{B}{>{\centering}p{0.022\textwidth}}
\newcolumntype{C}{>{\centering\arraybackslash}p{0.01\textwidth}}
\begin{table*}[t!]
  \caption{Comparison of mAP, the number of backward and forward passes, and FPS between baselines and our model on COCO $\rightarrow$ COCO-C. Our model consistently outperforms baselines on the two different backbones. Furthermore, Ours-Skip with ResNet notably reduces backward passes by as much as 90.5\%, leading to a significantly improved frames per second (FPS) rate by up to 109.9\%.}
  \vspace{-3mm}
  \scriptsize
  \centering
  \begin{tabularx}{1.0\textwidth}{p{0.07\textwidth} p{0.082\textwidth} |AAAAAAAAAAAAAAAA|A|BB|C}
    \toprule
    && \multicolumn{3}{c}{Noise} & \multicolumn{4}{c}{Blur} & \multicolumn{4}{c}{Weather} & \multicolumn{4}{c}{Digital} &\multicolumn{1}{c}{}&&\multicolumn{2}{c}{\# step}&\\
    \cmidrule(lr){3-5} \cmidrule(lr){6-9} \cmidrule(lr){10-13} \cmidrule(lr){14-17} \cmidrule(lr){20-21}
    Backbone & Method & Gau & Sht & Imp & Def & Gls & Mtn & Zm & Snw & Frs & Fog & Brt & Cnt & Els & Px & Jpg & \underline{Org.} & Avg. & For.&Back.&FPS\\
    \midrule
    \multirow{5}{5em}{Swin-T~\cite{liu2021swin}} 
    & Direct-Test & 9.7 & 11.4 & 10.0 & 13.4 & 7.5 & 12.1 & 5.2 & 20.7 & 24.8 & 36.1 & 36.0 & 12.9 & 19.1 & 4.9 & 15.8 & 43.0 & 17.7 & 80K &0&21.5\\
    & ActMAD & 10.7 & 12.0 & 9.4 & 12.3 & 5.7 & 9.5 & 4.5 & 15.3 & 17.5 & 27.6 & 28.2 & 1.1 & 16.7 & 2.6 & 8.7 & 36.3  & 13.9 &80K&80K&8.3\\
    & Mean-Teacher &10.0&12.1&11.2&12.8&8.1&12.1&4.9&19.6&23.7&34.9&34.0&8.0&18.9&6.1&17.6&41.0&17.2&160K&80K&6.9\\
    &\CCG Ours &\CCG \textbf{13.6} &\CCG \textbf{16.6} &\CCG \textbf{16.1} &\CCG \textbf{14.0} &\CCG \textbf{13.6} &\CCG \textbf{14.2} & \CCG\textbf{8.3} &\CCG\textbf{23.7}&\CCG \textbf{27.2} &\CCG\textbf{37.4} &\CCG 3\textbf{6.4}&\CCG \textbf{27.2} &\CCG\textbf{27.2} &\CCG22.2&\CCG \textbf{22.3} &\CCG \textbf{42.3} &\CCG \textbf{22.6} &\CCG80K&\CCG80K&\CCG9.5\\
    &\CCG Ours-Skip  &\CCG 13.3 &1\CCG5.3 &\CCG15.1 &\CCG 14.0 &\CCG12.8 &\CCG13.9 &\CCG6.5 &\CCG22.0&\CCG 25.4 &\CCG35.5 &\CCG34.9 &\CCG26.5 &\CCG25.9&\CCG \textbf{23.4} &\CCG20.2&\CCG 41.2&\CCG 21.6&\CCG80K&\CCG9.7K&\CCG17.7\\
    \midrule 
    \multirow{7}{5em}{ResNet50~\cite{he2016deep}} 
    & Direct-Test & 9.1 & 11.0 & 9.8 & 12.6 & 4.5 & 8.8 & 4.6 & 19.1 & 23.1 & 38.4 & 38.0 & 21.4 & 15.6 & 5.3 & 11.9 & 44.2 & 17.3 &80K&0&25.8\\
    & NORM & 9.9 & 11.9 & 11.0 & 12.6 & 5.2 & 9.1 & 5.1 & 19.4 & 23.5 & 38.2 & 37.6 & 22.4 & 17.2 & 5.7 & 10.3 & 43.4 & 17.5 &80K&0&25.8\\
    & DUA & 9.8 & 11.7 & 10.8 & 12.8 & 5.2 & 8.9 & 5.1 & 19.3 & 23.7 &38.4 & 37.8 & 22.3 & 17.2 & 5.4 & 10.1 & \textbf{44.1} & 17.1 &80K&0&25.8\\
    & ActMAD & 9.1 & 9.6 & 7.0 & 11.0 & 3.2& 6.1 & 3.3& 12.8 &14.0& 27.7& 27.8 &3.9& 12.9 &2.3& 7.2 &34.3 & 10.5 &80K&80K&9.6\\
    & Mean-Teacher &9.6&12.5&12.0&4.0&2.9&4.8&3.1&16.2&23.5&35.1&34.0&21.8&16.6&8.2&12.7&40.3&14.5&160K&80K&8.1\\
    &\CCG Ours & \CCG12.7&\CCG\textbf{17.8}&\CCG\textbf{17.5}&\CCG12.4&\CCG11.5&	\CCG11.3&\CCG\textbf{6.6}&\CCG\textbf{22.8}&\CCG\textbf{26.9}&\CCG\textbf{38.6}&\CCG\textbf{38.5}&\CCG\textbf{28.0}&\CCG\textbf{25.1}&\CCG21.2&\CCG\textbf{22.2}&\CCG41.8&\CCG\textbf{22.2}&\CCG80K&\CCG80K&\CCG10.1\\ 
    &\CCG Ours-Skip&\CCG\textbf{14.4}&\CCG17.1&\CCG16.0&\CCG\textbf{13.9}&\CCG\textbf{11.7}&\CCG\textbf{12.2}&\CCG6.3&\CCG22.1&\CCG25.5&\CCG37.7&\CCG37.1&\CCG25.5&\CCG24.1&\CCG\textbf{23.1}&\CCG21.1&\CCG42.8&\CCG21.9&\CCG80K&\CCG7.6K&\CCG21.2\\
    \bottomrule
  \end{tabularx}
  \label{tab:coco}
  \vspace{-2mm}
\end{table*}

\noindent \textbf{COCO $\rightarrow$ COCO-C} simulates continuous and drastic realistic test domain changes over a long sequence. MS-COCO \cite{coco} collects 80 classes of common objects in their natural context with 118k training images and 5k validation images. COCO-C is created by employing 15 types of realistic corruptions \cite{corruption}, such as image distortion and various weather conditions, to simulate test domain changes. In the experiments, the model is only trained on the COCO train set and sequentially evaluated on each corruption in the COCO-C validation set during test-time for reproducing continually changing test domains. Finally, the model is evaluated on the original COCO validation set to assess how well it preserves knowledge of the original domain (denoted as Org.).

\noindent \textbf{SHIFT-(Discrete / Continuous)} \cite{shift} is a synthetic driving image dataset with 6 classes under different conditions using five weather attributes (\textit{clear, cloudy, overcast, fog, rain}) and three time-of-day attributes (\textit{daytime, dawn, night}). 
In \textbf{SHIFT-Discrete}, there are image sets for each attribute, and \nj{the model is sequentially evaluated} on these attributes, \textit{cloudy} $\rightarrow$ \textit{overcast} $\rightarrow$ \textit{foggy} $\rightarrow$ \textit{rainy} $\rightarrow$ \textit{dawn} $\rightarrow$ \textit{night} $\rightarrow$ \textit{clear} \nj{which contains} 2.4k, 1.6k, 2.7k, 3.2k, 1.2k, 1.4k, and 2.8k validation images, respectively. This simulates scenarios where the domain undergoes drastic \nj{changes}. In \textbf{SHIFT-Continuous}, \nj{the model is evaluated} on four sequences, each consisting of 4k frames, continuously transitioning from \textit{clear} to \textit{foggy} (or \textit{rainy}) and back to \textit{clear}.


\subsection{Implementation Detail}
\label{sec:implementation}
We experiment with Faster-RCNN~\cite{ren2016faster_rcnn} models using ResNet50~\cite{he2016deep} and Swin-Tiny~\cite{liu2021swin} as a backbone with FPN~\cite{lin2017feature}. For the COCO $\rightarrow$ COCO-C adaptation, we employ the publicity available models trained on COCO released in ~\cite{xu2021end} and ~\cite{liu2021swin} for ResNet5\nj{-} and Swin-Tiny\nj{-}based Faster-RCNN, respectively. For SHIFT experiments, models are trained on the training domain using the detectron2 framework following \cite{ren2016faster_rcnn} and \cite{liu2021swin}.
For test-time adaptation, \nj{we always set the learning} to 0.001 for the SGD optimizer, and $\alpha$ of Eq.~\ref{eq:ema_mean} to 0.01, while $\tau_1$ and $\tau_2$ are set to 1.1 and 1.05, respectively. We use the same hyper-parameters across all backbones and datasets. All experiments are conducted with a batch size of 4.

\subsection{Baselines}
\label{sec:baselines}
\textbf{Direct-Test} evaluates the model trained in the training domain without adaptation to the test domain. \textbf{ActMAD}~\cite{ActMAD} is a TTA method aligning the distribution of output features across all BN layers. To apply ActMAD to the Swin Transformer\nj{-}based model, we align the output features of the LN layers. We implement \textbf{Mean-Teacher} using a teacher-student network framework to reproduce as close as possible to TeST~\cite{TeST}, as its implementation is not publicly available. We follow the FixMatch~\cite{sohn2020fixmatch} augmentation method and report results after tuning all hyper-parameters in our scenario. \textbf{NORM}~\cite{NORM} and \textbf{DUA}~\cite{DUA}, TTA methods initially designed for classification, are directly applicable to detection tasks by either mixing a certain amount of current batch statistics or updating batch statistics via EMA. However, these are only compatible with architectures containing BN layers. Additional details are provided in \nj{A}ppendix.

\newcolumntype{A}{>{\centering}p{0.02\textwidth}}
\newcolumntype{B}{>{\centering}p{0.02\textwidth}}
\newcolumntype{C}{>{\centering}p{0.06\textwidth}}
\newcolumntype{D}{>{\centering}p{0.023\textwidth}}
\newcolumntype{E}{>{\centering\arraybackslash}p{0.02\textwidth}}
\begin{table*}
    \caption{Comparison of mAP, the number of backward and forward passes, and FPS between baselines and our model on SHIFT-Discrete and SHIFT-Continuous. Baselines perform effectively in a particular setting but lack generalizability across various settings. Our method consistently achieves results that are either better or on par with the best model in all settings, demonstrating its strong stability. Ours-Skip also effectively reduces the number of backward passes \nj{without compromising mAP performance}, resulting in a higher FPS.}
    \vspace{-3mm}
  \scriptsize
  \begin{tabularx}{1.0\textwidth}{p{0.07\textwidth} p{0.082\textwidth}|AAAAAAA|A|DD|A|CC|DD|E}
    \toprule
    && \multicolumn{11}{c}{SHIFT-Discrete} & \multicolumn{5}{c}{SHIFT-Continuous} \\\cmidrule(lr){3-13} \cmidrule(lr){14-18}
    &&\multicolumn{8}{c}{mAP}&\multicolumn{2}{c}{\# step}&&\multicolumn{2}{c}{mAP}&\multicolumn{2}{c}{\# Avg. step}&\\
    \cmidrule(lr){3-10}\cmidrule(lr){11-12}\cmidrule(lr){14-15}\cmidrule(lr){16-17}
Backbone & Method&cloudy&overc.&fog&rain&dawn&night&\underline{clear}&Avg. &For.&Back.&FPS&\underline{clear}$\leftrightarrow$fog&\underline{clear}$\leftrightarrow$rain&For.&Back.&FPS\\
    \midrule
        \multirow{5}{5em}{Swin-T~\cite{liu2021swin}} 
        & Direct-Test &50.0&38.9&23.1&45.1&26.9&39.5&45.9&38.5&15.3K&0&27.5&18.1&21.1&4K&0&28.3\\
        & ActMAD &49.8&38.4&21.4&43.1&19.0&32.0&44.8&35.5&15.3K&15.3K&9.3&15.6&16.3&4K&4K&9.8\\
        &Mean-Teacher&50.0&39.2&25.7&45.4&26.0&37.5&42.2&38.0&15.3K&15.3K&7.8&20.4&\textbf{24.3}&8K&4K&6.5\\
        &\CCG Ours &\CCG\textbf{50.3}&\CCG39.2&\CCG\textbf{32.2}&\CCG46.7&\CCG\textbf{30.4}&\CCG39.9&\CCG44.3&\CCG40.4&\CCG15.3K&\CCG15.3K&\CCG11.2&\CCG23.9&\CCG22.6&\CCG4K&\CCG4K&\CCG11.6\\
        &\CCG Ours-Skip &\CCG\textbf{50.3}&\CCG\textbf{39.7}&\CCG29.1&\CCG\textbf{47.1}&\CCG30.2&\CCG\textbf{41.5}&\CCG\textbf{45.9}&\CCG\textbf{40.6}&\CCG15.3K&\CCG6.1K&\CCG20.0&\CCG\textbf{25.1}&\CCG23.8&\CCG4K&\CCG0.83K&\CCG19.2\\ 
    \midrule 
        \multirow{7}{5em}{ResNet50~\cite{he2016deep}} 
        & Direct-Test & 49.4 & 37.9 & 19.7 & 43.1 & 20.1 & 35.3 & 45.6 & 35.9 & 15.3K& 0&30.1&12.1& 15.4 &4K&0&30.0\\
        & NORM &\textbf{49.7}&38.6&22.9&44.7&25.1&37.4&\textbf{45.5}&37.7&15.3K&0&30.1&16.9&19.4&4K&0&30.0\\
        & DUA &45.2&31.5&\textbf{27.7}&31.9&15.2&18.6&21.1&27.3&15.3K&0&30.1&\textbf{22.5}&22.4&4K&0&30.0\\
        & ActMAD &49.2&37.7&18.0&40.6&16.0&32.9&44.3&34.1&15.3K&15.3K&11.3&12.7 & 16.3&4K&4K&11.2\\
        &Mean-Teacher&49.6&38.4&26.8&43.4&26.6&33.1&41.6&37.1&15.3K&15.3K&9.9&16.0&20.8&8K&4K&9.8\\
        &\CCG Ours &\CCG\textbf{49.7}&\CCG38.7&\CCG27.4&\CCG\textbf{46.3}&\CCG27.4&\CCG37.6&\CCG43.8&\CCG38.7&\CCG15.3K&\CCG15.3K&\CCG12.9&\CCG20.9&\CCG21.9&\CCG4K&\CCG4K&\CCG13.9\\ 
        &\CCG Ours-Skip &\CCG\textbf{49.7}&\CCG\textbf{38.8}&\CCG26.9&\CCG46.2&\CCG\textbf{27.6}&\CCG\textbf{38.8}&\CCG45.0&\CCG\textbf{39.0}&\CCG15.3K&\CCG8.9K&\CCG21.5&\CCG20.0&\CCG\textbf{22.5}&\CCG4K&\CCG0.75K&\CCG21.3\\
    \bottomrule
  \end{tabularx}
   \label{tab:shift}
  \vspace{-2mm}
\end{table*}
\subsection{Main Results}
\label{sec:results}

We compare the performance of each method using mAP and efficiency metrics, including the number of forward and backward passes, as well as FPS during test-time adaptation. Results of COCO and SHIFT are in Tab.~\ref{tab:coco} and~\ref{tab:shift}, respectively.

\noindent\textbf{COCO $\rightarrow$ COCO-C.} 
\nj{Tab.~\ref{tab:coco} demonstrates the effective adaptation performance of \textit{Ours}} in the challenging COCO benchmark with 80 classes due to object-level class-wise feature alignment. \textit{ActMAD} also aligns feature distribution for TTA, but is not effective since it only aligns whole feature maps without considering specific classes \nj{in the image}. \textit{NORM} and \textit{DUA}, applicable only to ResNet~\cite{he2016deep}, show minimal performance improvement by adaptation as they are not specifically tailored for object detection and only modify batch statistics across the entire feature map. Additionally, \textit{ActMAD} and \textit{Mean-Teacher}, updating full parameters, gradually lose task knowledge in the continually changing test distributions, resulting in much lower performance on \underline{Org.}, the domain identical to the training data, than that of \textit{Direct-Test}. In contrast, \textit{Ours} effectively prevents catastrophic forgetting by freezing the original parameters of the models and updating only the adaptor, obtaining performance on par with \textit{Direct-Test} on the \underline{Org.} domain and consistently high performance across corrupted domains, with an average mAP improvement of 4.9\%p compared to that of \textit{Direct-Test}. Furthermore, leveraging the rapid adaptation ability of the adaptor, \textit{Ours-Skip}, which skips unnecessary adaptation, allows using only a maximum of about 12\% of the total samples for adaptation without significant performance loss. This leads to a substantial improvement in inference speed, more than doubling compared to other TTA methods, reaching over 17.7 FPS.

\noindent\textbf{SHIFT-Discrete.} \textit{Ours} is also effective in SHIFT, which simulates continuous changes in weather and time in driving scenarios according to the left section of Tab.~\ref{tab:shift}. Especially, \textit{Ours} shows significant improvements in mAP by 7-9\%p, particularly for the foggy and dawn attributes where \textit{Direct-Test} obtains lower performance due to severe domain shift. In contrast, with ActMAD, catastrophic forgetting takes place when adapting to the cloudy and overcast \nj{weather}. This \nj{is due to the updating of the full parameters, despite that \textit{Direct-Test} already shows proficient performance in these conditions}.  
As a result, the performance in the later domains is worse than that of the \textit{Direct-Test}. \textit{DUA}, which updates batch statistics using EMA, shows a gradual decrease in performance as the domain continuously changes, resulting in much lower performance in the original clear domain (\ie, \underline{clear}). On the other hand, \textit{NORM}, which utilizes the statistics of the current batch samples, exhibits no catastrophic forgetting and relatively good adaptation, as SHIFT is a relatively easier task compared to COCO due to having only 6 classes. Compared to \textit{NORM}, \textit{Ours} shows better adaptation performance, and is also applicable to \nj{BN-layer-free Swin Transformers.} 

\noindent\textbf{SHIFT-Continuous.} In scenarios where the test domain gradually changes across the entire sequence, \textit{Ours} also demonstrates effectiveness, improving mAP by up to 7\%p, as shown in the right section of Tab.~\ref{tab:shift}. While \textit{DUA} performs well in the clear to foggy transition, it is prone to catastrophic forgetting in situations where the sequence becomes longer, and the test domain changes more diversely, as seen in the left section. Our strategy for determining when model adaptation is necessary is particularly effective in SHIFT. It improves FPS by about 9, reaching about 20 FPS, while enhancing mAP. This is likely due to avoiding overfitting that can occur when adapting to all repetitive frames in SHIFT, which consists of continuous frames, leading to improvements in both inference speed and adaptation performance.

\subsection{Additional Analyses}
\label{sec:ablation_study}
We aim to demonstrate the effectiveness and detailed analysis of our proposed model in terms of 1) which parts of, 2) how, and 3) when the model should be updated.

\newcolumntype{A}{>{\centering}p{0.015\textwidth}}
\newcolumntype{B}{>{\centering}p{0.029\textwidth}}
\newcolumntype{D}{>{\centering}p{0.02\textwidth}}
\newcolumntype{C}{>{\centering\arraybackslash}p{0.015\textwidth}}
\begin{table}
    \caption{Comparison of adaptation performance (mAP)\nj{, the} number of trainable parameters (\# Params), and memory usage (Cache) according to which part of the backbone \nj{is} updated. SD / SC \nj{denotes} SHIFT-Discrete/Continuous, respectively.}
  \vspace{-3mm}
    \footnotesize
  \begin{tabularx}{0.48\textwidth}{p{0.055\textwidth}p{0.115\textwidth}|AABDAC}
    \toprule
    &&\multicolumn{2}{c}{mAP}& \multicolumn{2}{c}{\# Params} & \multicolumn{2}{c}{Cache}\\
    \cmidrule(lr){3-4} \cmidrule(lr){5-6} \cmidrule(lr){7-8}
    Backbone&Trainable Params& SD & SC & Num & Ratio & Avg. & Max \\
    \midrule
        \multirow{3}{5em}{Swin-T} 
        & Full-params & 38.4 & 20.6& 27.7M & 100\% &0.86&11.0\\
        & LayerNorm &38.5&20.0& 0.03M & 0.1\% &0.65&7.49\\
        &\CCG adaptor (Ours) &\CCG  40.4 &\CCG 23.2 &\CCG 0.15M &\CCG 0.5\% &\CCG 0.65&\CCG 6.96\\
    \midrule 
        \multirow{3}{5em}{ResNet50} 
        & Full-params & 37.6 &20.4& 23.7M & 100\% &1.65&9.29\\
        & BatchNorm & 37.9& 20.2&0.05M & 0.2\%&1.47&9.11\\
        & \CCG adaptor (Ours) &\CCG 38.7&\CCG 21.7&\CCG 0.21M &\CCG 0.9\% \CCG  &\CCG 1.48&\CCG 5.41\\
    \bottomrule
  \end{tabularx}
  \label{tab:abl_what}
  \vspace{-2mm}
\end{table}

\newcolumntype{A}{>{\centering}p{0.02\textwidth}}
\newcolumntype{B}{>{\centering}p{0.12\textwidth}}
\newcolumntype{C}{>{\centering}p{0.065\textwidth}}
\newcolumntype{D}{>{\centering\arraybackslash}p{0.065\textwidth}}
\begin{table}
  \caption{Ablation on each component of our loss. SHIFT-D / C denotes SHIFT-Discrete / Continuous, respectively. The left and right value in each cell corresponds to the mAP for the Swin-T and ResNet50 backbone, respectively.}
    \vspace{-3mm}
    \footnotesize
  \centering
  \begin{tabularx}{0.45\textwidth}{AB|CCD}
    \toprule
    $L_{img}$&$L_{obj}$&COCO&SHIFT-D.&SHIFT-C.\\
    \midrule
        -&-&17.7/\ 17.3&38.5/\ 35.9&19.6/\ 13.8\\
        \ding{52}&-& 16.7/\ 18.1&36.6/\ 37.0&19.1/\ 16.0\\
        \ding{52}&no class weight& 17.8/\ 18.9\ &39.7/\ 38.0&25.1/\ 23.4\\
        \ding{52}&class weight $w^{k,t}$&22.6/\ 22.2&40.4/\ 38.7&23.2/\ 21.7\\
    \bottomrule
  \end{tabularx}
  \label{tab:abl_how}
  \vspace{-2mm}
\end{table}

\noindent\textbf{\nj{Which part to update? }} 
Tab.~\ref{tab:abl_what} shows how updating different parts of the backbone model affects the performance and the memory usage during continual test-time adaptation. We compare \nj{(1)} updating full parameters, \nj{(2)} affine parameters of the normalization layer, and \nj{(3)} our proposed adaptor for each backbone on the SHIFT dataset. Although our adaptor has fewer parameters, about 0.9\% or less of the full parameters, it demonstrates the best adaptation performance. Updating only the affine parameters of the normalization layer, while having fewer parameters, seems less effective for adaptation in object detection compared to classification~\cite{TENT,EATA}. Additionally, our adaptor requires only about 60\% of the memory compared to updating the full parameters, making it memory-efficient.

\noindent\textbf{Ablation study on each component in our loss.}
Tab.~\ref{tab:abl_how} presents the effects of image-level feature alignment, $L_{img}$, object-level feature class-wise alignment $L_{obj}$, and class frequency weighting $w^{k,t}$ proposed to address class imbalance. Aligning only the image-level feature distribution with $L_{img}$ (first row) leads to modest adaptation in the ResNet50 backbone, while performance in the Swin-T backbone is even lower than without adaptation. Notably, aligning object-level features with $L_{obj}$ leads to a substantial improvement, with the mAP increasing by approximately 10\%p  compared to the no-adaptation scenario. Introducing class-specific frequency-based weighting $w^{k,t}$, despite a slight performance decrease in the SHIFT-Continuous setting, proves highly effective, particularly in scenarios with significant class imbalance, such as COCO with 80 classes, where it enhances the mAP by around 5\%p.

\begin{figure}
\begin{subfigure}{.5\linewidth}
  \centering
  \includegraphics[width=1.0\linewidth]{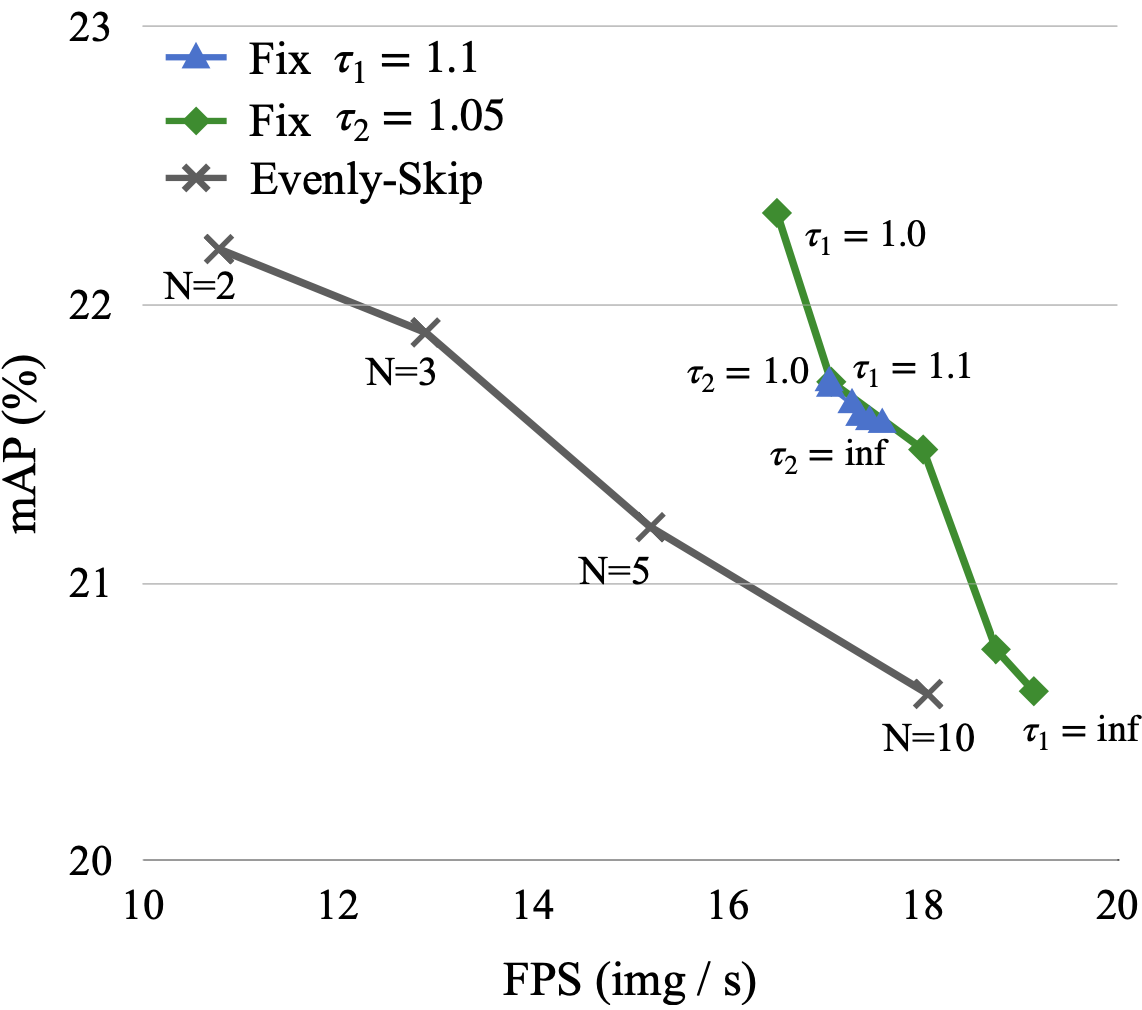}
  \caption{Swin Transformer backbone}
  \label{fig:abl_when_swint}
\end{subfigure}%
\begin{subfigure}{.5\linewidth}
  \centering
  \includegraphics[width=1.0\linewidth]{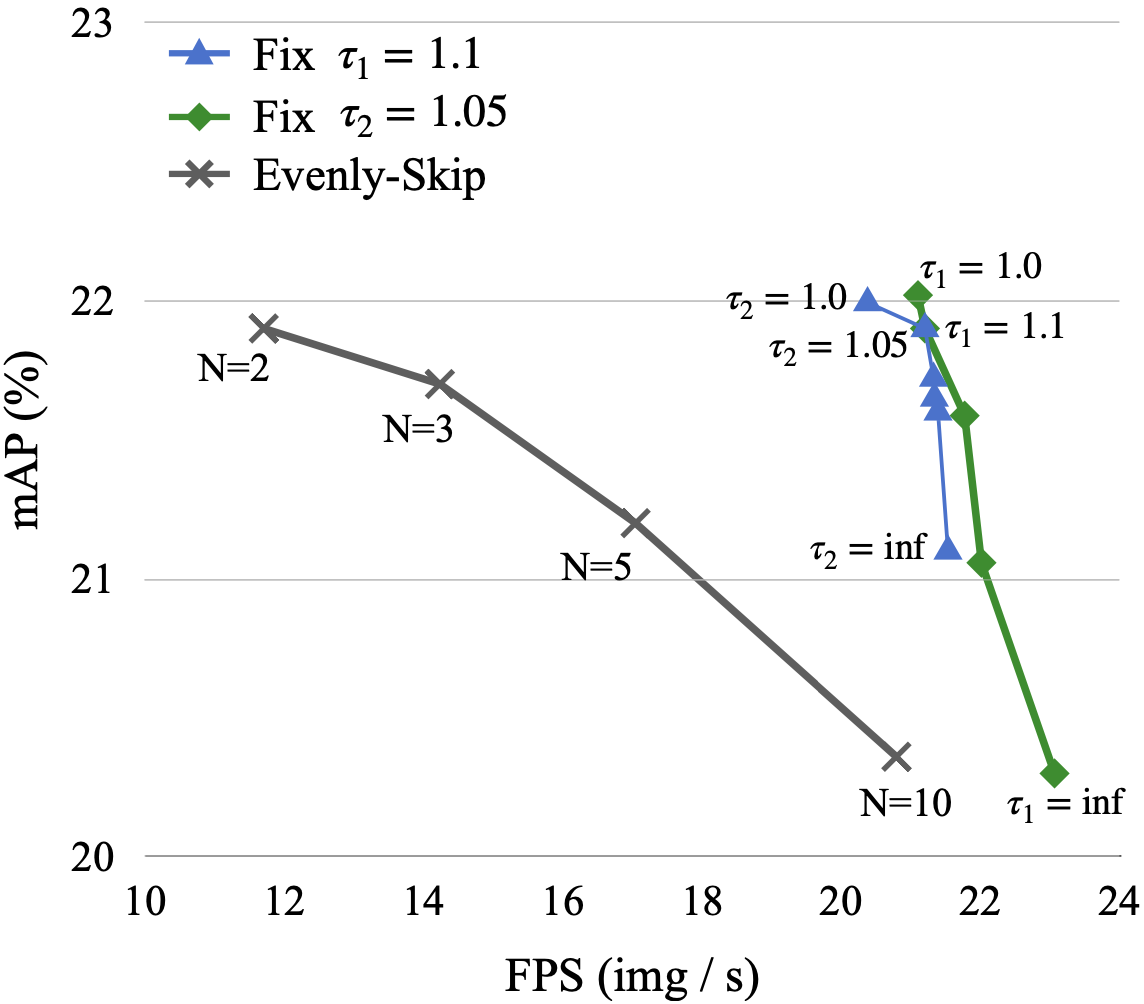}
  \caption{ResNet50 backbone}
  \label{fig:abl_when_r50}
\end{subfigure}
\vspace{-4mm}
\caption{
Comparison of mAP and FPS from \textit{Ours-Skip}  with varying values of $\tau_1$ (\textcolor{ForestGreen}{$\blacklozenge$}) and $\tau_2$ (\textcolor{RoyalBlue}{$\blacktriangle$}) against \textit{Evenly-Skip} ($\times$), adapting every $N$-th instances, on COCO$\rightarrow$COCO-C using both (a) Swin-T and (b) ResNet50. The upward and rightward movement indicates a better strategy with higher mAP and faster inference speed, showing that \textit{Ours-Skip} is consistently better than \textit{Evenly-Skip}.
}
\label{fig:abl_when}
\vspace{-4mm}
\end{figure}

\noindent\textbf{\nj{Trade-off} between adaptation performance and efficiency according to \nj{different} skipping strategies.}
Fig.~\ref{fig:abl_when} presents mAP and FPS depending on the values of \nj{$\tau_1$ and $\tau_2$} in the Sec.~\ref{subsec:when} on COCO $\rightarrow$ COCO-C, which are used for two criteria to determine when the adaptation is needed. We also show the simple baseline \textit{Evenly-Skip}, which adapts every $N$-th step and skips the rest. In Fig.~\ref{fig:abl_when}, the blue lines (\textcolor{RoyalBlue}{$\blacktriangle$}) 
show the results when $\tau_1$ is changing from 1.0 to infinity, where only criterion 2 is used, while $\tau_2$ is fixed at 1.05. As $\tau_1$ decreases, more adaptation is required, leading to slower FPS but higher mAP. The green lines (\textcolor{ForestGreen}{$\blacklozenge$}) show the results of changing \nj{$\tau_2$}, where `$\tau_2=\text{inf}$' denotes using only \nj{criterion 1}, without criterion 2. 
For all main experiments, we set $\tau_1$ and $\tau_2$ as 1.1 and 1.05, respectively, considering the balance between mAP and FPS. Additionally, our skipping strategy consistently outperforms \textit{Evenly-Skip}, achieving higher values in both mAP and FPS. 
This indicates that our criterion for deciding when to bypass model updates provides an effective balance \nj{between accuracy and speed}.

\begin{figure}
\begin{subfigure}[h]{.5\textwidth}
  \centering
  \includegraphics[width=.9\linewidth]{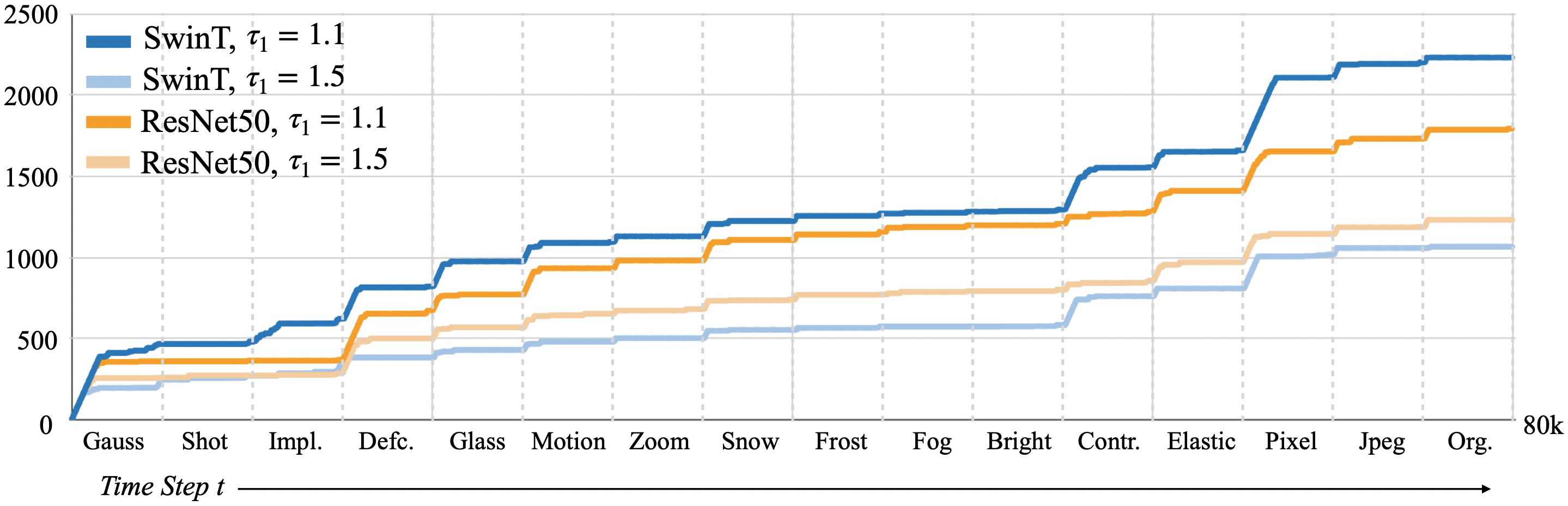}
  \caption{Accumulated number of backward steps}
  \label{fig:analysis1a}
\end{subfigure}%
\hfill
\begin{subfigure}[h]{.5\textwidth}
  \centering
  \includegraphics[width=.9\linewidth]{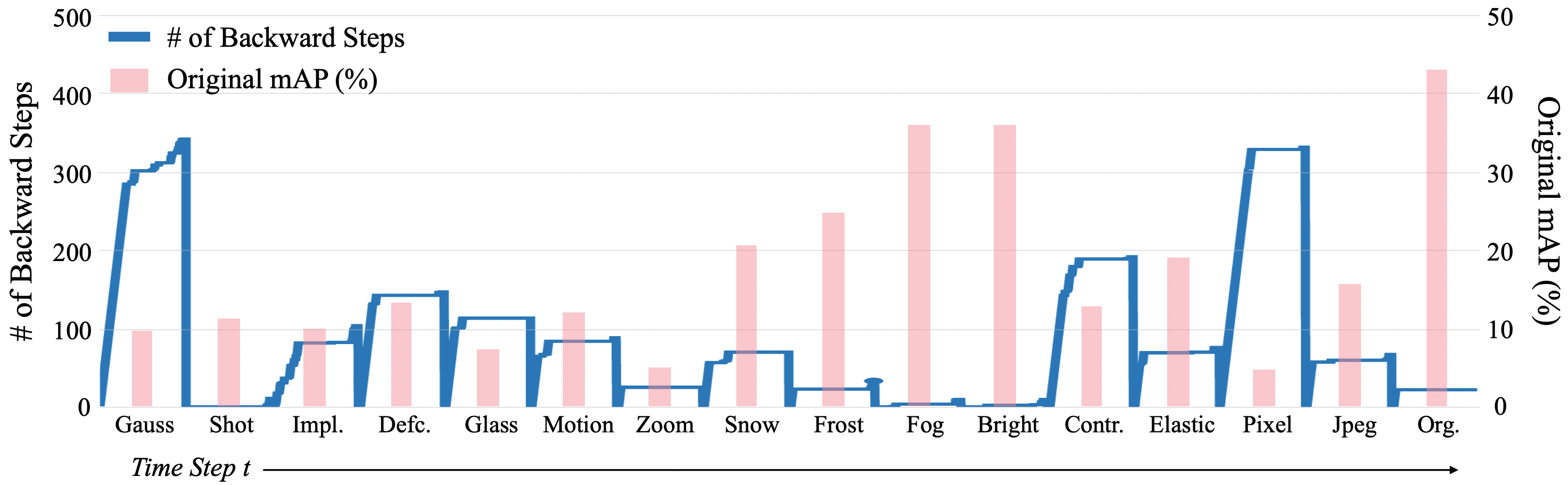}
  \caption{Number of backward steps and mAP of Direct-Test in each domain}
  \label{fig:analysis1b}
\end{subfigure}
\vspace{-1mm}
\caption{Analysis of the adaptation of \textit{Ours-Skip}.}
\label{fig:analysis1}
\vspace{-2mm}
\end{figure}

\noindent\textbf{\nj{When do models actually update?}} We analyze when the model actually skips adaptation and only performs inference or actively utilizes test samples for model adaptation based on the two criteria we propose. This analysis is conducted in COCO to COCO-C with 15 corruption domains and 1 original domain. Fig.~\ref{fig:analysis1a} plots the number of backward passes, \ie, the number of batches of test samples used for adaptation, with different values of $\tau_1$ for the two backbones. The horizontal and vertical axes represent sequentially incoming test domains and the cumulative backward numbers, respectively. A steep slope in a region indicates frequent adaptation, while a gentle slope indicates \nj{skipping adaptation, performing only inference.} Notably, even without explicit information about when the test domain changes, the model actively performs adaptation, especially \nj{right after the test domain changes.} 
This trend is consistent regardless of changes in $\tau$ value or backbone type. Furthermore, it is evident that the number of backward passes is primarily determined by \nj{the value of} $\tau_1$ rather than the type of backbone, suggesting that a consistent $\tau_1$ \nj{value} can be used irrespective of the backbone.
Fig.~\ref{fig:analysis1b} visually represents the adaptation tendencies by dividing backward steps for each domain in the case of Swin-T backbone with $\tau_1=1.1$. More clearly, it shows that adaptation occurs actively around the points where each domain changes, and afterward, adaptation happens intermittently or almost not at all. The light pink bars represent the performance of \textit{Direct-Test}, showing that domains with initially high model performance tend to have less adaptation, while domains with lower performance initially need more adaptation. In other words, the amount of skipping adaptation is proportional to the amount of the domain shift. Interestingly, the second domain, 'Shot Noise', shows almost no adaptation despite the lower performance of the \textit{Direct-Test}. We conjecture that the preceding domain, 'Gaussian Noise', shares a similar nature of noise, leading the model to decide that additional adaptation steps may not be necessary. As a result, our skipping strategy enables the model to efficiently adapt, considering both the original domain the model is trained on and the previous domain the model has been adapted to.
\section{Conclusion}

We introduce an efficient Continual Test-time Adaptation (CTA) method for object detection in the continually changing domain. Our approach involves 1) lightweight adaptors, 2) class-wise object-level feature alignment, and 3) skipping unnecessary adaptation. These contributions collectively yield a highly efficient and effective adaptation method, showcasing robustness to diverse domain shifts, and achieving notable improvements in mAP performance across various CTA scenarios \nj{without serious slowdown in the inference speed}.
\clearpage
\setcounter{page}{1}
\maketitlesupplementary



\section{Additional Details for Baselines}
\label{sec:Baseline}
We provide additional implementation details for each baseline model. Our framework incorporates all baseline models using the official code except \textit{Mean-Teacher}. The results of the experiments are reported based on the optimal hyperparameters that yield the best results in our scenario.

\noindent\textbf{ActMAD}~\cite{ActMAD} As ActMAD exclusively conducts experiments on the KITTI dataset, where all images have a constant height and width (e.g., 370 x 1224), ensuring consistent feature map sizes for all samples. ActMAD can easily align them along the spatial axis. However, in the general setting of object detection tasks, such as the COCO benchmark set, where image sizes and width-to-height ratios vary, aligning feature maps along the spatial axis becomes challenging due to different sizes. To adapt ActMAD to our COCO $\rightarrow$ COCO-C scenario, we perform center cropping on the feature maps to match the size of training domain feature maps and the current test sample feature maps. We employ a learning rate of 1e-5 for COCO and 1e-4 for SHIFT, respectively.

\noindent\textbf{Mean-Teacher}
As the official code of TeST~\cite{TeST} is not available, we implement the EMA-updated Teacher and Student models following TeST~\cite{TeST}, to conduct experiments in our scenarios. TeST involves three forward steps for a batch: forwarding weakly augmented samples through the student network, strong augmented samples through the teacher network, and original samples through the teacher network for outputs. However, for a fair comparison, we perform two forward steps, forwarding the original sample through the teacher network and strong augmented samples through the student network, to make predictions before adaptation for every samples. We utilize a learning rate of 1e-5 and set the EMA update rate for the teacher network to 0.999.

\noindent\textbf{NORM}~\cite{NORM} We set the hyperparameter $N$ that controls the trade-off between training statistics and estimated target statistics as 128.

\noindent\textbf{DUA}~\cite{DUA} We set the momentum decay as 0.94, minimum momentum constant as 1e-4, and the initial momentum decay as 1e-3. 

\newcolumntype{A}{>{\centering}p{0.03\textwidth}}
\newcolumntype{B}{>{\centering}p{0.022\textwidth}}
\newcolumntype{D}{>{\centering}p{0.0\textwidth}}
\newcolumntype{C}{>{\centering\arraybackslash}p{0.02\textwidth}}
\begin{table}
    \caption{Comparison of adaptation performance (mAP), the number of trainable parameters (\# Params), and memory usage (Cache) according to $r$ of Sec.~\ref{subsec:adaptor}, the bottleneck reduction ratio in the adaptor. We set $r$ as 32 for all our experiments in the main paper. SD / SC \nj{denotes} SHIFT-Discrete / Continuous, respectively.}
  \vspace{-3mm}
    \footnotesize
  \begin{tabularx}{0.48\textwidth}{p{0.055\textwidth}p{0.02\textwidth}|ABBAABC}
    \toprule
    &&\multicolumn{3}{c}{mAP}& \multicolumn{2}{c}{\# Params} & \multicolumn{2}{c}{Cache}\\
    \cmidrule(lr){3-5} \cmidrule(lr){6-7} \cmidrule(lr){8-9}
    Backbone&$r$& COCO &SD & SC & Num & Ratio & Avg. & Max \\
    \midrule
        \multirow{7}{5em}{Swin-T} 
        & 1 & 22.6& 40.0 & 21.3& 4.33M & 15.7\% &0.75&7.51\\
        & 2 & 22.6&40.3&23.2& 2.17M & 7.85\% &0.73&7.27\\
        & 4 & 22.6&40.4&23.2& 1.09M & 3.95\% &0.70&7.06\\
        & 8 & 22.6&40.4&23.2& 0.55M & 2.00\% &0.69&7.00\\
        & 16 & 22.6&40.4&23.2& 0.28M & 1.02\% &0.67&6.98\\
        &\CCG 32 & \CCG 22.6&\CCG  40.4 &\CCG 23.2 &\CCG 0.15M &\CCG 0.54\% &\CCG 0.65&\CCG 6.96\\
        & 64 & 22.6&40.4&23.2& 0.08M & 0.29\% &0.65&6.95\\
    \midrule 
        \multirow{7}{5em}{ResNet50} 
        & 1 &22.5& 38.7 & 20.8& 6.31M & 26.7\% &1.55&5.89\\
        & 2 &22.4&38.7&20.9& 3.16M & 13.4\% &1.51&5.64\\
        & 4 &22.3&38.6&21.3& 1.59M & 6.71\% &1.49&5.52\\
        & 8 &22.3&38.6&21.4& 0.80M & 3.39\% &1.48&5.46\\
        & 16 &22.2&38.6&21.4& 0.41M & 1.73\% &1.48&5.43\\
        &\CCG 32 &\CCG 22.2&\CCG  38.7 &\CCG 21.4 &\CCG 0.21M &\CCG 0.89\% &\CCG 1.48&\CCG 5.41\\
        & 64 &22.1&38.7&21.3& 0.11M & 0.48\% &1.48&5.40\\
    \bottomrule
  \end{tabularx}
  \label{tab:abl_bottleneck_size}
  \vspace{-2mm}
\end{table}

\newcolumntype{A}{>{\centering}p{0.025\textwidth}}
\newcolumntype{B}{>{\centering}p{0.067\textwidth}}
\newcolumntype{C}{>{\centering}p{0.01\textwidth}}
\newcolumntype{D}{>{\centering\arraybackslash}p{0.02\textwidth}}
\begin{table*}[t!]
  \caption{Comparison of mAP, the number of backward and forward passes, FPS, and memory usage between baselines and our models on the continually changing KITTI datasets (\textit{Fog} $\rightarrow$ \textit{Rain} $\rightarrow$ \textit{Snow} $\rightarrow$ \textit{Clear}). Our models improve mAP@50 by 15.1 and 11.3 for Swin-T and ResNet50 backbone, respectively, compared to \textit{Direct-Test} while maintaining comparable FPS. All experiments are conducted with a batch size of 16.}
  \vspace{-3mm}
  \scriptsize
  \centering
  \begin{tabularx}{0.97\textwidth}{p{0.08\textwidth} p{0.1\textwidth} |AAAA|A|BAAAA|A|A|AD}\\
    \toprule
    && \multicolumn{5}{c}{mAP@50} & \# For. Steps &\multicolumn{5}{c}{\# Backward Steps}& FPS &\multicolumn{2}{c}{Cache}\\
    \cmidrule(lr){3-7} \cmidrule(lr){9-13} \cmidrule(lr){15-16}
    Backbone & Method & Fog & Rain & Snow & \underline{Clear} & Avg. & All &Fog & Rain & Snow & \underline{Clear} & All & Avg. &Avg. & Max\\
    \midrule
\multirow{5}{5em}{Swin-T}  & Direct-Test & 46.9 & 69.5 & 28.7 & 89.6 & 58.7 & 936 & 0 & 0 & 0 & 0 & 0 & 24.7 & 0.4 & 5.5 \\
 & ActMAD & 53.3 & 78.1 & 41.2 & 90.7 & 65.8 & 936 & 234 & 234 & 234 & 234 & 936 & 16.8 & 0.8 & 21.9 \\
 & Mean-Teacher & 54.5 & 80.2 & 43.2 & 92.4 & 67.6 & 936 & 234 & 234 & 234 & 234 & 936 & 10.0 & 1.0 & 22.6 \\
 &\CCG Ours &\CCG 56.7 &\CCG 82.1 &\CCG 64.6 &\CCG 91.8 &\CCG 73.8 &\CCG 936 &\CCG 234 &\CCG 234 &\CCG 234 &\CCG 234 &\CCG 936 &\CCG 17.1 &\CCG 0.4 &\CCG 11.8 \\
 &\CCG Ours-Skip &\CCG 57.4 &\CCG 81.5 &\CCG 64.3 &\CCG 91.3 &\CCG 73.6 &\CCG 936 &\CCG 234 &\CCG 65 &\CCG 224 &\CCG 36 &\CCG 559 &\CCG 22.9 &\CCG 0.4 &\CCG 11.8 \\
    \midrule
\multirow{7}{5em}{ResNet50}  & Direct-Test & 33.4 & 63.5 & 29.8 & 88.6 & 53.8 & 936 & 0 & 0 & 0 & 0 & 0 & 27.7 & 0.8 & 4.3 \\
 & NORM & 38.4 & 66.4 & 35.9 & 87.3 & 57.0 & 936 & 0 & 0 & 0 & 0 & 0 & 27.7 & 0.8 & 4.3 \\
 & DUA & 34.8 & 67.7 & 30.9 & 89.0 & 55.6 & 936 & 0 & 0 & 0 & 0 & 0 & 27.7 & 0.8 & 4.3 \\
 & ActMAD & 40.4 & 66.5 & 42.7 & 84.5 & 58.5 & 936 & 234 & 234 & 234 & 234 & 936 & 18.5 & 1.6 & 22.6 \\
 & Mean-Teacher & 39.6 & 71.3 & 43.5 & 88.2 & 60.6 & 936 & 234 & 234 & 234 & 234 & 936 & 11.1 & 1.8 & 31.1 \\
 & \CCG Ours & \CCG 45.6 & \CCG 71.4 & \CCG 52.5 & \CCG 88.3 & \CCG 64.5 & \CCG 936 & \CCG 234 & \CCG 234 & \CCG 234 & \CCG 234 & \CCG 936 & \CCG 18.8 & \CCG 0.8 & \CCG 9.4 \\
 & \CCG Ours-Skip & \CCG 45.8 & \CCG 71.3 & \CCG 50.9 & \CCG 88.4 & \CCG 64.1 & \CCG 936 & \CCG 234 & \CCG 111 & \CCG 98 & \CCG 45 & \CCG 488 & \CCG 24.5 & \CCG 0.8 & \CCG 9.4 \\
    \bottomrule
  \end{tabularx}
  \label{tab:kitti}
  \vspace{-2mm}
\end{table*}

\begin{figure*}[t]
\begin{subfigure}[t]{1.0\textwidth}
  \centering
  \includegraphics[width=1.0\linewidth]{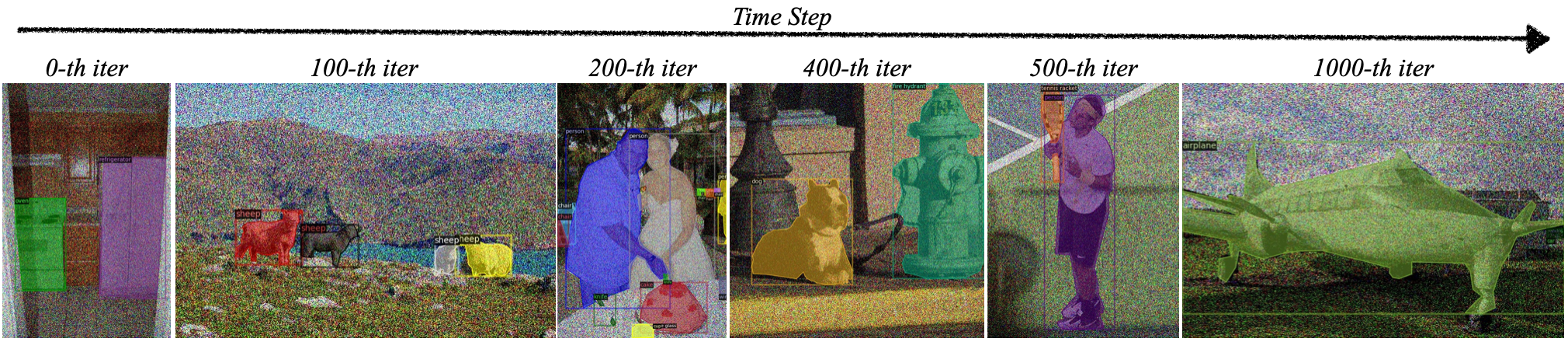}
  \caption{GT bounding boxes.}
  \label{fig:qual1_gt}
\end{subfigure}%
\hfill
\begin{subfigure}[t]{1.0\textwidth}
  \centering
  \includegraphics[width=1.0\linewidth]{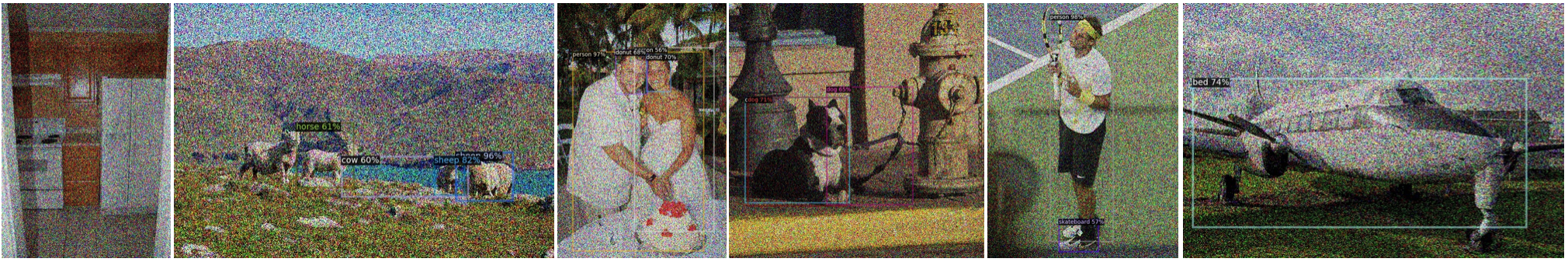}
  \caption{Prediction results of \textit{Direct-Test}.}
  \label{fig:qual1_direct_test}
\end{subfigure}
\hfill
\begin{subfigure}[t]{1.0\textwidth}
  \centering
  \includegraphics[width=1.0\linewidth]{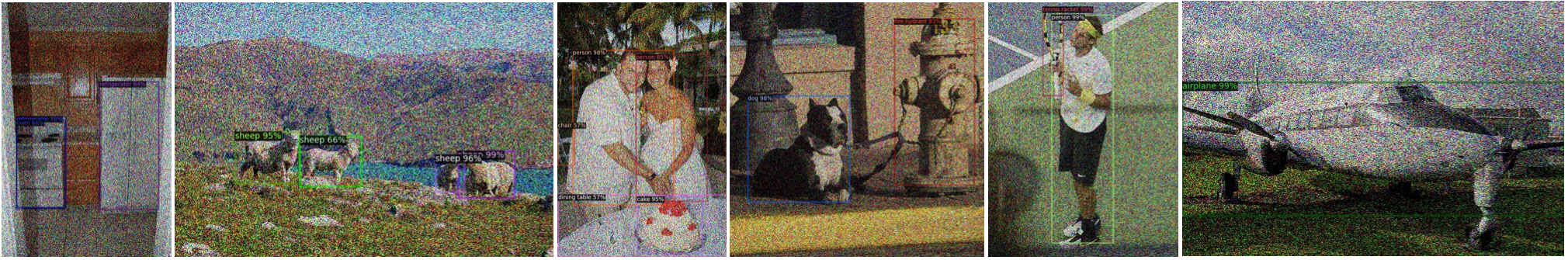}
  \caption{Prediction results of \textit{Ours}.}
  \label{fig:qual1_ours}
\end{subfigure}
\vspace{-1mm}
\caption{Results of COCO images corrupted by \textit{Shot-Noise}. In the analysis of Sec.~\ref{sec:ablation_study}, we conjecture that \textit{Ours} largely skips adaptation in \textit{Shot-Noise} domain, despite the low mAP of \textit{Direct-Test}, because the model has already adapted to a similar domain, \textit{Gaussian-Noise}. In (c), at the first step before adaptation to the \textit{Shot-Noise}, our model already predicts 'Oven' and 'Refrigerator' which \textit{Direct-Test} fails to detect. This results in a much faster adaptation, and \textit{Ours} successfully detects various objects, including rare ones such as 'Fire Hydrants', in the remaining images of the \textit{Shot-Noise} domain.}
\label{fig:qual1}
\vspace{-2mm}
\end{figure*}

\begin{figure*}[t]
\begin{subfigure}[t]{1.0\textwidth}
  \centering
  \includegraphics[width=1.0\linewidth]{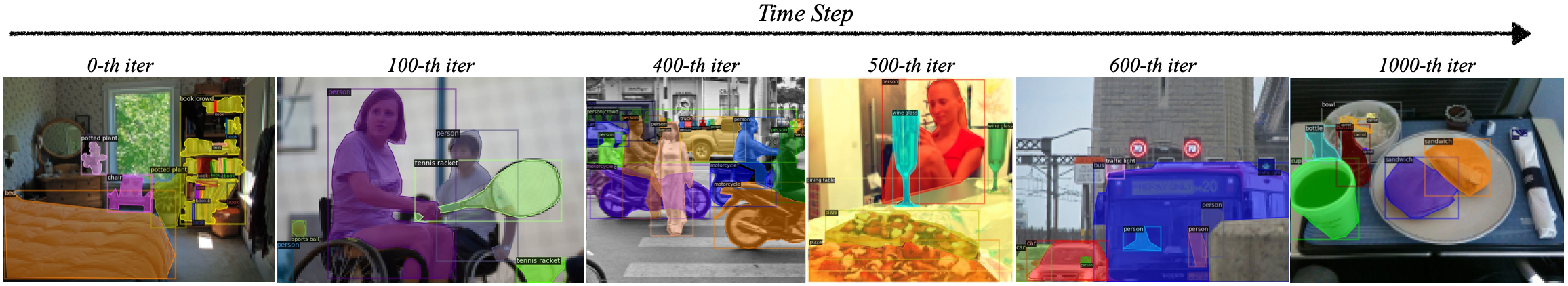}
  \caption{GT bounding boxes.}
  \label{fig:qual2_gt}
\end{subfigure}%
\hfill
\begin{subfigure}[t]{1.0\textwidth}
  \centering
  \includegraphics[width=1.0\linewidth]{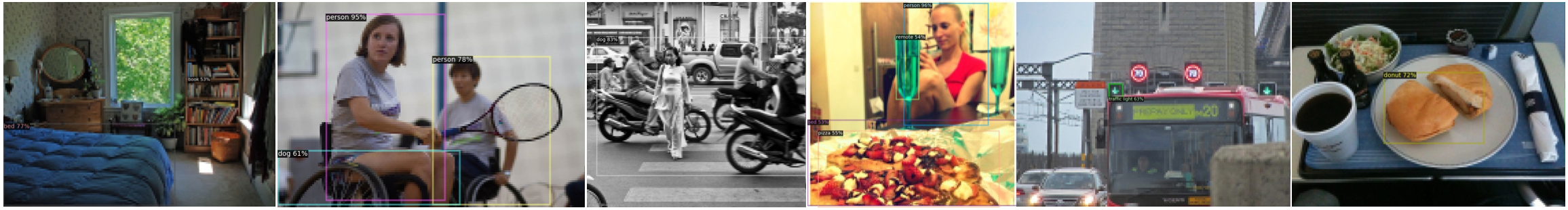}
  \caption{Prediction results of \textit{Direct-Test}.}
  \label{fig:qual2_direct_test}
\end{subfigure}
\hfill
\begin{subfigure}[t]{1.0\textwidth}
  \centering
  \includegraphics[width=1.0\linewidth]{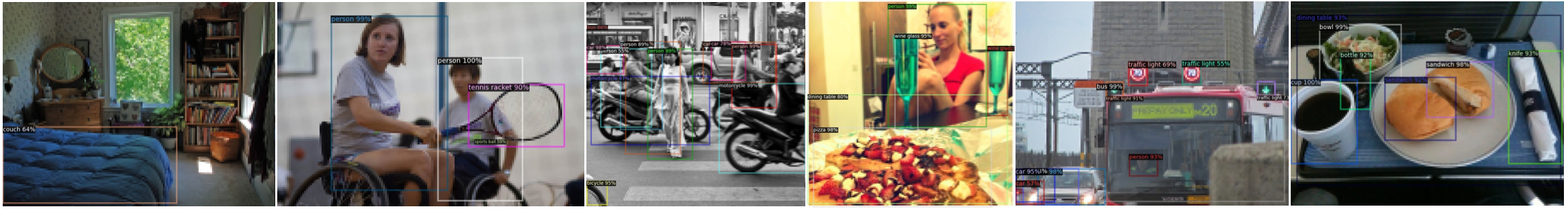}
  \caption{Prediction results of \textit{Ours}.}
  \label{fig:qual2_ours}
\end{subfigure}
\vspace{-1mm}
\caption{Results for COCO images corrupted by \textit{Pixelate}. In the \textit{Pixelate} domain, where the model has already experienced various corruptions in a long sequence, \textit{Ours} initially incorrectly detects objects. In (c), it misidentifies a bed as a couch in the first step. However, it rapidly adapts to the \textit{Pixelate} domain and effectively detects various objects. Notably, even in cases where \textit{Direct-Test} correctly identifies objects but with low confidence, \textit{Ours} detects them with much higher confidence.}
\label{fig:qual2}
\vspace{-2mm}
\end{figure*}

\begin{figure*}[t]
\begin{subfigure}[t]{1.0\textwidth}
  \centering
  \includegraphics[width=1.0\linewidth]{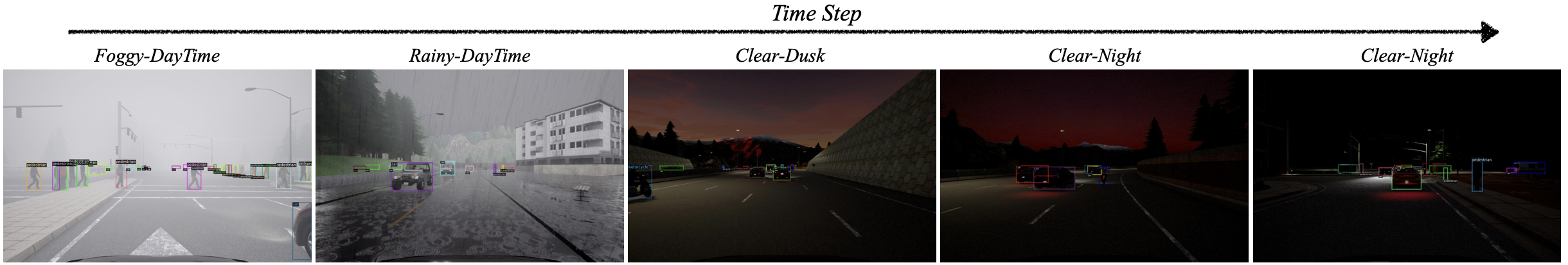}
  \caption{GT bounding boxes.}
  \label{fig:qual3_gt}
\end{subfigure}%
\hfill
\begin{subfigure}[t]{1.0\textwidth}
  \centering
  \includegraphics[width=1.0\linewidth]{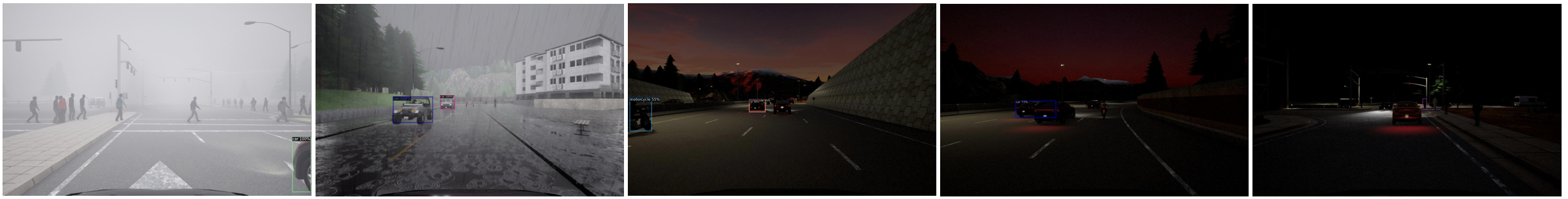}
  \caption{Prediction results of \textit{Direct-Test}.}
  \label{fig:qual3_direct_test}
\end{subfigure}
\hfill
\begin{subfigure}[t]{1.0\textwidth}
  \centering
  \includegraphics[width=1.0\linewidth]{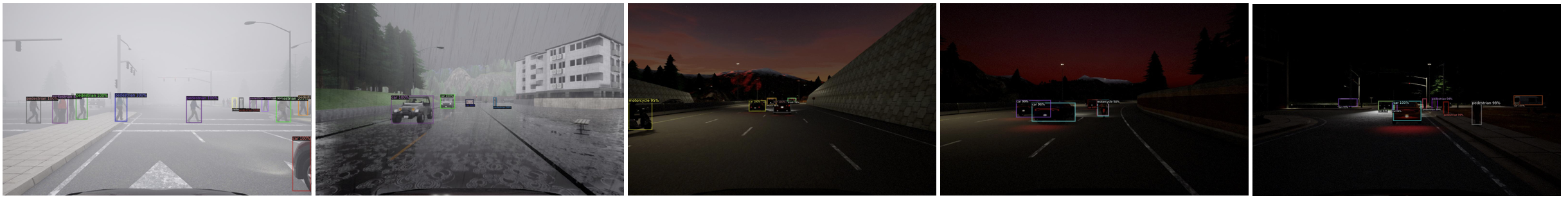}
  \caption{Prediction results of \textit{Ours}.}
  \label{fig:qual3_ours}
\end{subfigure}
\vspace{-1mm}
\caption{Results for SHIFT-Discrete with continually changing attributes, \textit{foggy} $\rightarrow$ \textit{rainy} $\rightarrow$ \textit{dawn} $\rightarrow$ \textit{night}.}
\label{fig:qual3}
\vspace{-2mm}
\end{figure*}

\section{The effect of Bottleneck Reduction Ratio in the Adaptor}
\label{sec:abl_bottleneck}
Table~\ref{tab:abl_bottleneck_size} shows the results for COCO $\rightarrow$ COCO-C, SHIFT-Discrete, and SHIFT-Continuous based on the dimension reduction ratio ($r$) discussed in Section~\ref{subsec:adaptor}, representing the ratio of bottleneck size compared to the input size in the adaptor. The adaptation performance remains consistent across different $r$ values. However, in the case of $r=1$ in SHIFT experiments, mAP decreases, potentially due to catastrophic forgetting resulting from a large number of adaptable parameters. Since increasing the value of $r$ significantly reduces the number of learnable parameters and memory usage, we set $r$ to 32 in all other experiments.


\section{Results on the KITTI Dataset}
We conduct additional experiments on the KITTI~\cite{geiger2013vision} dataset, the commonly used object detection dataset consisting of driving scenes with 8 classes (car, van, truck, person, person sitting, cyclist, tram, misc). To simulate the continually changing domains, we use the following scenario (\textit{Fog} $\rightarrow$ \textit{Rain} $\rightarrow$ \textit{Snow} $\rightarrow$ \textit{Clear}) as done in~\cite{ActMAD}. We use the physics-based rendered dataset~\cite{halder2019physics} for \textit{fog} and \textit{rain} and simulate \textit{snow} using the corruption library from~\cite{hendrycks2019benchmarking}. We use the same split of \cite{ActMAD}, which divides the 7,441 training samples into 3,740 training and 3,741 test samples. We train the Faster-RCNN using 3,741 training samples representing the \textit{Clear} attribute with Swin-Transformer and ResNet50 backbones, and evaluate it sequentially on \textit{Fog}, \textit{Rain}, \textit{Snow}, and \textit{Clear} test samples.

We conduct all experiments with a batch size of 16 on 1 RTX A6000 GPU. Table~\ref{tab:kitti} shows the mAP@50, the number of forward and backward steps, FPS, and memory usage (Cache). \textit{Ours} improves the mAP@50 by 15.1 and 10.7 for Swin-T and ResNet50 backbones, respectively, compared to \textit{Direct-Test}. Compared to \textit{ActMAD} and \textit{Mean-Teacher}, our model not only improves the adaptation performance but also reduces memory usage, as we update only an extremely small number of parameters of the adaptor. Furthermore, using our skipping criteria of Sec.~\ref{subsec:when} with $\tau=1.1$ and $\beta=1.05$, we can improve FPS by more than 5.8 without sacrificing mAP@50, resulting in much faster inference speed compared to other TTA baselines.

\section{Qualitative Results}
Fig.~\ref{fig:qual1} and \ref{fig:qual2} and Fig.~\ref{fig:qual3} show the qualitative results of \textit{Ours} and \textit{Direct-Test} which predict the samples without adaptation for COCO $\rightarrow$ COCO-C and SHIFT, respectively.

\subsection{COCO $\rightarrow$ COCO-C}
Fig.~\ref{fig:qual1} and \ref{fig:qual2} compare the prediction results for COCO images corrupted. When the model encounters test images with various corruptions sequentially (\textit{Gaussian-Noise} $\rightarrow$ \textit{Shot-Noise} $\rightarrow$ \textit{Impulse-Noise} $\rightarrow$ \textit{Defocus-Blur} $\rightarrow$ \textit{Glass-Blur} $\rightarrow$ \textit{Motion-Blur} $\rightarrow$ \textit{Zoom-Blur} $\rightarrow$ \textit{Snow} $\rightarrow$ \textit{Frost} $\rightarrow$ \textit{Fog} $\rightarrow$ \textit{Brightness} $\rightarrow$ \textit{Contrast} $\rightarrow$ \textit{Elastic-Transform} $\rightarrow$ \textit{Pixelate} $\rightarrow$ \textit{JPEG-Compression} $\rightarrow$ \textit{Original}), Fig.~\ref{fig:qual1} and \ref{fig:qual2} shows the results when the test images are corrupted by \textit{Shot-Noise} and \textit{Pixelate}, respectively. Compared to \textit{Direct-Test}, our model adapts to the current domain within a few steps, such as 100 iterations, and detects various objects very well in the remaining incoming images. 

\subsection{SHIFT-Discrete}
Fig.~\ref{fig:qual3} shows the qualitative results for SHIFT-Discrete. In the SHIFT-Discrete scenario, the model encounters environments sequentially, transitioning from \textit{cloudy} $\rightarrow$ \textit{overcast} $\rightarrow$ \textit{foggy} $\rightarrow$ \textit{rainy} $\rightarrow$ \textit{dawn} $\rightarrow$ \textit{night} $\rightarrow$ \textit{clear}. Figure.~\ref{fig:qual3} selectively shows the \textit{foggy} $\rightarrow$ \textit{rainy} $\rightarrow$ \textit{dawn} $\rightarrow$ \textit{night} sequence, where the domain gap from the original \textit{clear} environments is relatively large. 
Compared to \textit{Direct-Test}, \textit{Ours} detects various objects such as 'cars' and 'pedestrians' regardless of distribution changes.

%
%
{
    \small
    \bibliographystyle{ieeenat_fullname}
    \bibliography{main}
}


\end{document}